\documentclass[preprint,12pt]{elsarticle}

\usepackage{amsfonts}
\usepackage[ruled,vlined]{algorithm2e}

\usepackage{array}
\usepackage[caption=false,font=normalsize,labelfont=sf,textfont=sf]{subfig}
\usepackage{textcomp}
\usepackage{stfloats}
\usepackage{url}
\usepackage{verbatim}
\usepackage{graphicx}

\usepackage{amsmath}
\usepackage{amssymb}
\usepackage{booktabs}
\usepackage{multirow}
\usepackage[normalem]{ulem}
\usepackage{chngcntr}
\usepackage{xcolor}
\usepackage{pifont}
\usepackage{tabularx}
\usepackage{geometry}
\usepackage{placeins}

\usepackage{makecell}
\usepackage{adjustbox}

\usepackage{hyperref}
\usepackage{float} 

\useunder{\uline}{\ul}{}

\DeclareMathOperator*{\argmin}{arg\,min}
\usepackage[capitalize]{cleveref}
\crefname{section}{Section}{Sections}
\crefname{table}{Table}{Tables}
\crefname{figure}{Figure}{Figures}
\crefname{equation}{Eq.}{Eqs.}
\crefname{appendix}{Appendix.}{Appendix.}

\journal{Pattern Recognition}

\begin{document}

\begin{frontmatter}

\title{Multi-class Image Anomaly Detection for Practical Applications: Requirements and Robust Solutions}

\author[label1]{Jaehyuk~Heo}
\ead{jaehyuk.heo@snu.ac.kr}
\author[label1]{Pilsung~Kang\corref{cor1}}
\ead{pilsung\_kang@snu.ac.kr}
\affiliation[label1]{organization={Department of Industrial Engineering, Seoul National University},
            country={Republic of Korea}}

\cortext[cor1]{Corresponding author}

%% Abstract
\begin{abstract}
Recent advances in image anomaly detection have extended unsupervised learning-based models from single-class settings to multi-class frameworks, aiming to improve efficiency in training time and model storage. When a single model is trained to handle multiple classes, it often underperforms compared to class-specific models in terms of per-class detection accuracy. Accordingly, previous studies have primarily focused on narrowing this performance gap. However, the way class information is used, or not used, remains a relatively understudied factor that could influence how detection thresholds are defined in multi-class image anomaly detection. These thresholds, whether class-specific or class-agnostic, significantly affect detection outcomes. In this study, we identify and formalize the requirements that a multi-class image anomaly detection model must satisfy under different conditions, depending on whether class labels are available during training and evaluation. We then re-examine existing methods under these criteria. To meet these challenges, we propose Hierarchical Coreset (HierCore), a novel framework designed to satisfy all defined requirements. HierCore operates effectively even without class labels, leveraging a hierarchical memory bank to estimate class-wise decision criteria for anomaly detection. We empirically validate the applicability and robustness of existing methods and HierCore under four distinct scenarios, determined by the presence or absence of class labels in the training and evaluation phases. The experimental results demonstrate that HierCore consistently meets all requirements and maintains strong, stable performance across all settings, highlighting its practical potential for real-world multi-class anomaly detection tasks.
\end{abstract}

%% Keywords

\end{frontmatter}

\section{Introduction}
\label{sec:intro}

Image anomaly detection is a computer vision task that involves not only determining whether an image contains an abnormal region but also localizing such regions within the image. Unlike conventional classification tasks, it requires both detection and localisation capabilities. However, the development of effective anomaly detection models is hindered by the limited availability and high variability of abnormal data, which makes supervised learning approaches impractical in many cases. To address this, recent studies have primarily adopted an unsupervised learning paradigm, where models are trained solely on normal data and detect anomalies based on deviations from the learned patterns of normality. This unsupervised image anomaly detection (UIAD) approach has demonstrated strong performance and has been widely applied in domains such as visual inspection~\cite{visual_inspection}, medical imaging~\cite{medical1, medical2}, and video surveillance systems~\cite{surveillance1, surveillance2, surveillance3}.

Recently, the UIAD framework has evolved from the conventional one-class setting, where a separate model is trained for each class, to a multi-class setting, in which a single model handles multiple classes simultaneously~\cite{diad,vitad,invad,mambaad,mint_ad,uniad}. One-class UIAD (OC-UIAD) suffers from a major scalability issue, as model storage and computation increase linearly with the number of classes. To improve scalability, multi-class UIAD (MC-UIAD) approaches have been proposed, aiming to detect anomalies across multiple classes using a single model. However, MC-UIAD methods often show inferior performance compared to OC-UIAD, as the representation learned from aggregated data across different classes tends to be less discriminative. Prior studies have primarily focused on minimizing this performance gap~\cite{mint_ad}.

\begin{figure*}[t!]
\begin{center}
\includegraphics[width=1\linewidth]{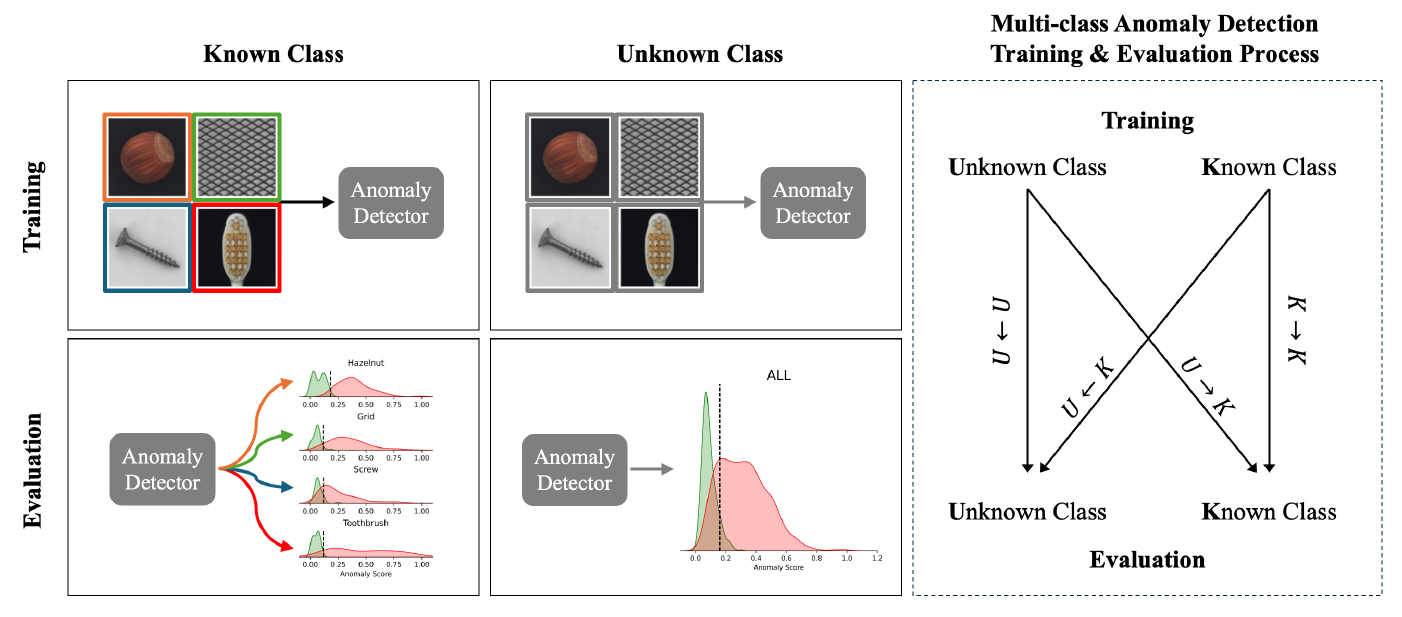}
\end{center}
\vspace{-20pt}
\caption{Multi-class image anomaly detection scenarios required for practical applications. The figure presents four combinations of training and evaluation conditions based on the availability of class information: (1) $K \rightarrow K$: training and evaluation are both conducted with known classes, (2) $U \rightarrow K$: the model is trained with unknown classes and evaluated on known classes, (3) $K \rightarrow U$: the model is trained with known classes and evaluated on unknown classes, and (4) $U \rightarrow U$: both training and evaluation are performed without class information. To be practically applicable, an anomaly detection model should be capable of handling all these cases.}
\label{fig:scenarios}
\end{figure*}

This study goes beyond the performance trade-off between OC-UIAD and MC-UIAD and instead aims to redefine the requirements of MC-UIAD from a practical deployment perspective. Most existing MC-UIAD research either does not use class labels explicitly during training or implicitly leverages them during evaluation. However, the presence or absence of class labels can significantly affect both the training methodology and the performance of multi class anomaly detection in real-world scenarios.

Figure~\ref{fig:scenarios} illustrates four possible scenarios for training and evaluation in MC-UIAD, depending on whether class labels are available. To enable applicability across all these scenarios, we define two key requirements for MC-UIAD:

\begin{itemize}
    \item \textbf{Requirement 1}: The model should be trainable regardless of whether class labels are available.
    \item \textbf{Requirement 2}: The model should maintain comparable anomaly detection performance during evaluation, irrespective of the availability of class labels.
\end{itemize}

When class labels are provided, the model can learn separate representations of normal data per class, enabling clearer boundaries between normal and anomalous patterns~\cite{mint_ad}. In contrast, when class labels are unavailable, the model must learn a shared embedding space for all data. While this may hinder class-wise separation, it has a practical advantage of eliminating annotation costs. Most existing MC-UIAD methods adopt label-agnostic training~\cite{diad,vitad,invad,mambaad,uniad}, but as a result, they lack the flexibility to adapt to label-available environments.

\begin{figure*}[t!]
\begin{center}
\includegraphics[width=1\linewidth]{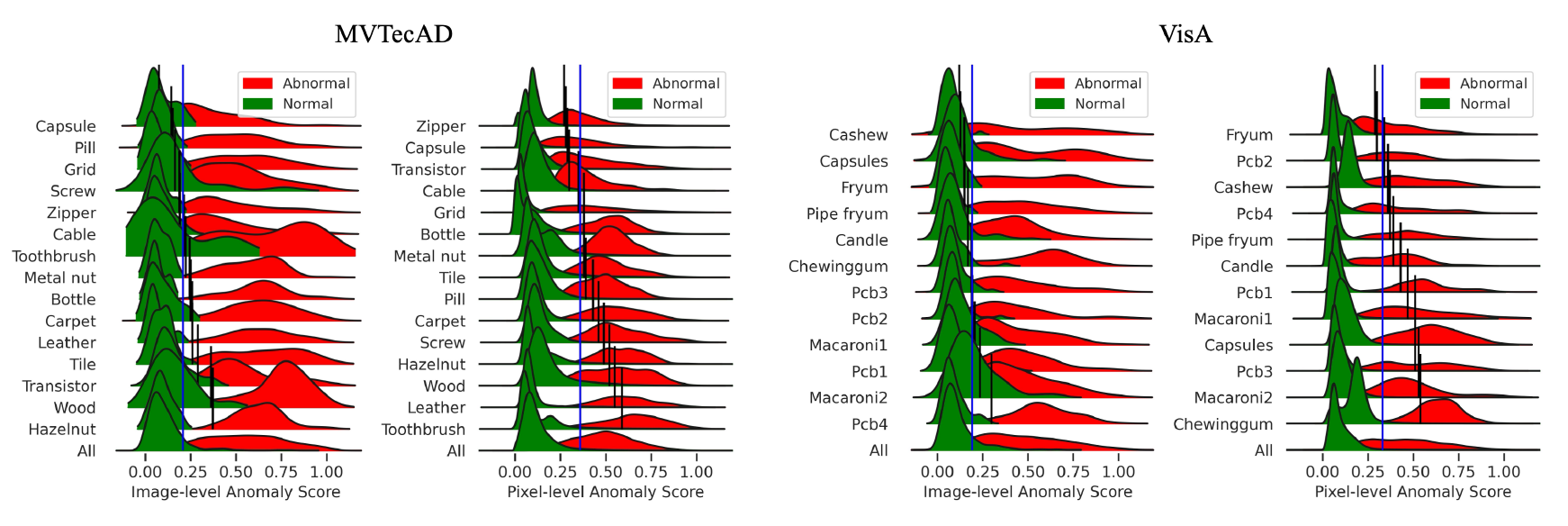}
\end{center}
\vspace{-20pt}
\caption{Class-wise and overall anomaly score distributions on MVTecAD and VisA datasets at both image- and pixel-levels. Red and green areas represent distributions of abnormal and normal samples, respectively. Black vertical lines indicate the optimal threshold determined per class, while the blue vertical line indicates the threshold optimized over all classes (i.e., all samples combined).}
\label{fig:thresholds}
\end{figure*}

Moreover, the presence or absence of class labels during evaluation critically impacts detection performance. Anomaly detection typically requires setting a decision threshold to distinguish between normal and abnormal instances. When class labels are available, optimal class-specific thresholds can be used to maximize detection accuracy. However, in the absence of such labels, a single global threshold must be applied across all classes. Due to inter-class variability, this often results in suboptimal performance for certain classes, manifesting as increased false positives or false negatives. Figure~\ref{fig:thresholds} demonstrates the variation in optimal thresholds at both the image- and pixel-level across classes. Although evaluation performance heavily depends on the presence of class labels, many existing works~\cite{uniad,rlr,uniformaly,vq_flow} evaluate models as if class labels were available during inference, even though they were not used during training. This discrepancy highlights the need for a systematic re-evaluation of whether a method can maintain robust performance without class information.

In both training and inference phases, the availability of class labels plays a significant role in MC-UIAD. However, acquiring such labels incurs high annotation costs and may even introduce noise if mislabeling occurs~\cite{noisy_label}. Therefore, to satisfy the above requirements, it is crucial to design a model capable of distinguishing data distributions and adapting thresholds at a class level without relying on explicit class labels.

To this end, we propose Hierarchical Coreset (\mbox{HierCore}), a novel MC-UIAD framework designed to satisfy both requirements under varying label availability conditions. \mbox{HierCore} operates effectively even without class labels by leveraging semantic information to group data and constructing a hierarchical memory bank using normal samples within each semantic cluster. Specifically, it performs semantic clustering to estimate latent class groupings and assigns cluster-specific keys to build corresponding memory banks. During inference, new inputs are matched to their most relevant semantic cluster, and anomalies are detected based on thresholds defined within each memory bank. We evaluate \mbox{HierCore} across four scenarios defined by the presence or absence of class labels during training and evaluation, and empirically verify its ability to meet all the proposed requirements.

The main contributions of this paper are as follows:

\begin{enumerate}
    \item We define two requirements that a model must satisfy to enable the practical application of multi-class unsupervised image anomaly detection, depending on whether class information is used during training and evaluation.
    \item We systematically re-evaluate existing MC-UIAD methods under four realistic scenarios to assess whether they meet the two defined requirements, using four industrial benchmark datasets.
    \item We propose a semantic-aware, hierarchical memory bank framework called \mbox{HierCore}, which remains effective across all scenarios, including those where class labels are unavailable, and achieves robust anomaly detection performance.
\end{enumerate}

The remainder of this paper is structured as follows. Section~\ref{sec:rel_works} reviews existing MC-UIAD approaches and highlights limitations in their training and evaluation strategies. Section~\ref{sec:pro_method} introduces the proposed \mbox{HierCore} framework and its key components. Section~\ref{sec:exp} presents the experimental setup and results across the four defined scenarios. Finally, Section~\ref{sec:conclusion} concludes the paper and discusses future directions.

\section{Related Works}
\label{sec:rel_works}

\subsection{Multi-class Image Anomaly Detection}

MC-UIAD has been developed to enable a single model to detect anomalies across multiple classes, with a focus on improving efficiency in terms of model size and computational cost. Most prior works addressing this problem have employed reconstruction-based approaches that integrate features of normal data from multiple classes into a single model~\cite{diad,vitad,invad,mambaad,mint_ad,uniad}. Reconstruction-based methods learn a model to reconstruct the input image and detect anomalies by calculating the reconstruction error, the difference between the input and its reconstructed version. These approaches assume that a model trained on normal data will fail to accurately reconstruct anomalous inputs. However, they suffer from the \textit{identical shortcut} problem~\cite{identical_shortcut}, where even anomalous regions can be reconstructed. This problem becomes more severe as the complexity of the normal data increases~\cite{uniad}. To overcome these limitations, several enhancements have been proposed. You et al.~\cite{uniad} introduced UniAD, a Transformer-based model that decomposes input images into patches and applies a neighbor-masked attention module to avoid simply copying neighboring information. Jiang et al.~\cite{mint_ad} addressed the lack of generalized reconstruction capability across classes by introducing an inter-class inference refinement method that explicitly leverages class labels. Lu et al.~\cite{hvqad} tackled the identical shortcut problem by adopting vector quantization to learn discrete representations of normal data.

An alternative to reconstruction-based methods is the memory bank-based approach~\cite{cfa,padim,patchcore}, which stores features extracted from normal data and compares them with those of new inputs to determine whether they are anomalous. Since these models do not attempt to reconstruct the input, they are not affected by the identical shortcut problem. However, memory usage and computation grow with the number of stored features, posing a scalability challenge.

Our proposed approach inherits the advantages of memory bank-based methods while introducing a hierarchical memory bank structure that improves computational efficiency. This hierarchical design allows us to address both performance and scalability challenges in real-world MC-UIAD scenarios.

\begin{table}
\caption{Overview of training and evaluation settings in recent multi-class anomaly detection models. “Unknown” and “Known” indicate whether class labels are assumed to be unavailable or available, respectively. A check mark ($\checkmark$) denotes that the model is trained or evaluated under the corresponding setting.}\label{tab:related_works}
\vspace{5pt}
\centering
\begin{adjustbox}{max width=\textwidth}
\begin{tabular}{@{}c|cc|cc@{}}
\toprule
                         & \multicolumn{2}{c|}{Training}                       & \multicolumn{2}{c}{Evaluation}     \\ \cmidrule(l){2-5} 
\multirow{-2}{*}{Models} & Unkwown                  & Known                    & Unknown & Known                    \\ \midrule
UniAD                    & $\checkmark$ &                          &         & $\checkmark$ \\
ViTAD                    & $\checkmark$ &                          &         & $\checkmark$ \\
InvAD                    & $\checkmark$ &                          &         & $\checkmark$ \\
MambaAD                  & $\checkmark$ &                          &         & $\checkmark$ \\
MINT-AD                  &                          & $\checkmark$ &         & $\checkmark$ \\
HierCore & $\checkmark$ & $\checkmark$ & $\checkmark$ & $\checkmark$ \\ \bottomrule
\end{tabular}
\end{adjustbox}
\end{table}

\subsection{Training and Evaluation of Multi-class Image Anomaly Detection}

Table~\ref{tab:related_works} categorizes existing MC-UIAD methods based on whether class information is used during training and evaluation. Most state-of-the-art methods, such as UniAD~\cite{uniad}, ViTAD~\cite{vitad}, InvAD~\cite{invad}, and MambaAD~\cite{diad}, are designed under the assumption of \textit{unknown-class training}, where class labels are not available during training. As a result, these methods do not leverage class-specific learning strategies. In contrast, MINT-AD~\cite{mint_ad} assumes \textit{known-class training} and adopts a class-aware architecture that learns representations separately for each class, improving inter-class discrimination. However, such approaches are difficult to apply when class labels are not available. Furthermore, most previous studies assume the availability of class labels during evaluation (\textit{known-class evaluation}), which limits their applicability to real-world scenarios where class labels are often unavailable. This gap highlights a lack of consideration for \textit{unknown-class evaluation} conditions, and the robustness of these models under such settings remains underexplored.

To address this gap, we argue that it is critical to examine whether existing methods can maintain performance under unknown evaluation conditions. We emphasise the importance of developing methods that operate effectively regardless of whether class labels are available during training and evaluation. We also propose \mbox{HierCore}, a novel framework designed to overcome these limitations. \mbox{HierCore} is capable of robust performance even without class information, thanks to its use of a semantic-aware hierarchical memory bank. This structure enables the model to cluster normal data semantically and define a separate anomaly detection threshold for each cluster, even in the absence of explicit class labels.

\begin{figure*}[t]
\begin{center}
\includegraphics[width=1\linewidth]{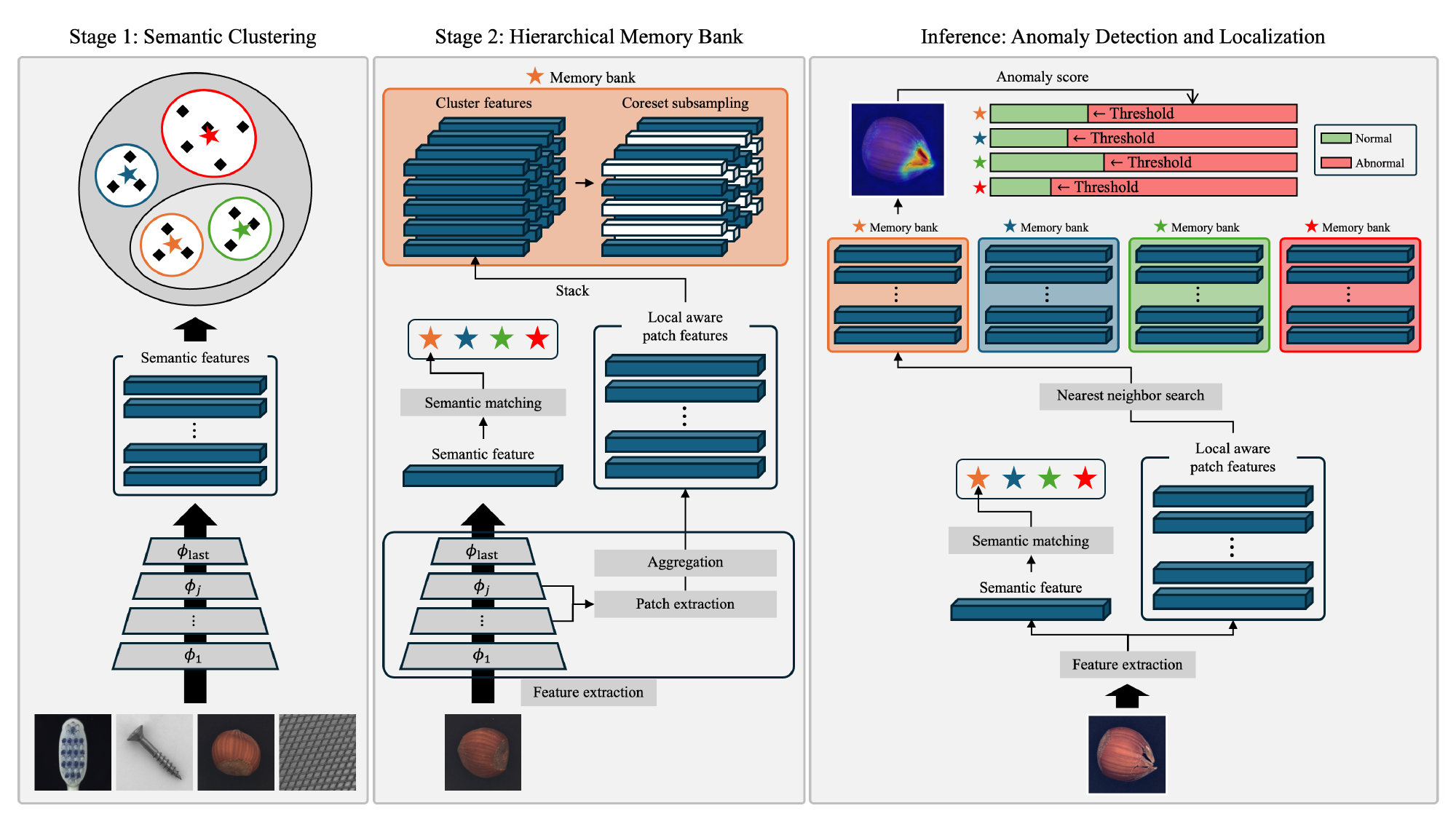}
\end{center}
\vspace{-20pt}
\caption{Overview of the HierCore. In Stage 1, semantic features extracted from normal images using a pre-trained encoder are clustered, and the cluster centroids are used as semantic keys for matching. In Stage 2, both semantic features and local patch features are extracted, and patch features are grouped based on the closest semantic key. A coreset-based memory bank is then constructed for each cluster. During inference, features from a test image are matched to the nearest semantic key, and anomaly scores are computed via nearest neighbor search within the corresponding memory bank. Cluster-specific thresholds are used to detect anomalies independent of class labels.}
\label{fig:overview}
\end{figure*}

\section{Proposed Method}
\label{sec:pro_method}

Figure~\ref{fig:overview} illustrates the overall architecture and inference procedure of the proposed \mbox{HierCore} framework. \mbox{HierCore} is designed to detect both the presence and location of anomalies without access to class labels. The framework first estimates the latent class of an input image based on its semantic features and then matches the image to the corresponding memory bank. Anomalies are detected by comparing the input features with stored normal patterns in that memory bank.

As shown in Figure~\ref{fig:overview}, \mbox{HierCore} operates in two main stages:

\begin{itemize}
    \item \textbf{Stage 1: Semantic Clustering.} Without access to ground-truth class labels, the semantic embedding of the input image is used to estimate its most likely class, and the image is assigned to the corresponding memory bank (Section~\ref{sec:semantic_clustering}).
    \item \textbf{Stage 2: Hierarchical Memory Bank.} During training, local features of normal images are stored in class-specific memory banks to capture representative normal patterns (Section~\ref{sec:memory_bank}). 
    \item \textbf{Inference: Anomaly Detection and Localization.} During inference, the estimated class and its corresponding memory bank are used to detect both anomalous regions and the overall anomaly score (Section~\ref{sec:anomaly_detection}).
\end{itemize}

By constructing a hierarchical memory bank structure based on pseudo-classes, \mbox{HierCore} significantly reduces computational cost compared to single-bank methods such as PatchCore. These efficiency gains are analysed in Section~\ref{sec:complexity}.

\subsection{Semantic Clustering}
\label{sec:semantic_clustering}

To enable class-agnostic operation, \mbox{HierCore} estimates pseudo-classes from semantic features rather than relying on explicit labels. This is achieved using a pre-trained image encoder $\phi$ (e.g., trained on ImageNet~\cite{imagenet}), which hierarchically extracts both low-level (edges, colors) and high-level (semantic) features from input images~\cite{deconvnet}. The final-layer output $\phi_{\text{last}}$ is especially effective at distinguishing images with different semantics within the normal set $\mathcal{X}_N = \{x \mid y=0\}$. We define the full set of semantic feature vectors $\textbf{e}$ as:

\[
\textbf{e} = \left\{ \phi_{\text{last}}(x_i) \,\middle|\, x_i \in \mathcal{X}_N \right\}.
\]

To estimate pseudo-classes, we adopt the FINCH algorithm~\cite{finch}, which clusters semantic embeddings without requiring a predefined number of clusters. FINCH is computationally efficient, insensitive to hyperparameters, and generates hierarchical clusters. Among the resulting cluster hierarchies, we select the level with the highest Silhouette score~\cite{silhouettes}, and define its cluster count as $K$.

\subsection{Hierarchical Memory Bank}
\label{sec:memory_bank}

The estimated semantic clusters effectively separate images with distinct patterns, allowing each group to reflect a unique normal distribution. To associate an image with its corresponding cluster, we define a representative key for each cluster as the mean vector of the semantic features contained within that cluster, denoted as $c = \{c_1, c_2, \dots, c_K\}$:

\begin{align}
c_k = \frac{1}{N_k}\sum_{i=1}^{N_k} e_i,
\end{align}

\noindent where $N_k$ is the number of normal images assigned to cluster $k$. Each semantic feature vector $e_i$ from a normal image is assigned to the closest cluster based on the L2 distance $d$ to each cluster's key. The corresponding cluster index $k_i^*$ is defined as follows:

\begin{align}
k^*_i = \argmin_{k \in \{1,2,\dots,K\}} d(e_i, c_k).
\end{align}

To detect localized anomalies, we extract local features from intermediate layers $\phi_j(x_i)$ of the backbone network, as in PatchCore~\cite{patchcore}. These features are divided into patch-wise representations using a patch extraction function $\mathcal{P}$, which splits an input feature map into local aware patch features. These are then used to construct a memory bank $\mathcal{M}_k$ for each cluster $k$:

\begin{align}
\mathcal{M}_k = \bigcup_{x_i \in \mathcal{X}_{N,k}} \mathcal{P}(\phi_j(x_i)),
\end{align}

\noindent where $\mathcal{X}_{N,k}$ is the set of normal images assigned to cluster $k$. Since local features from the same cluster may contain redundant information~\cite{patchcore,coreset}, we apply a coreset selection algorithm based on $k$-center clustering to reduce memory size. The final compressed memory bank $\mathcal{M}^*_k$ is defined as:

\begin{align}
\mathcal{M}_{k,c}^* = \argmin_{\mathcal{M}_{k,c} \subset \mathcal{M}_k} \max_{m \in \mathcal{M}_k} \min_{n \in \mathcal{M}_{k,c}} \| m - n \|_2,
\end{align}

\noindent ensuring both representativeness and efficiency.

\subsection{Anomaly Detection and Localization}
\label{sec:anomaly_detection}

To detect anomalies in a new image, \mbox{HierCore} performs the following steps: (1) Estimate the image's pseudo-class via semantic embedding; (2) Extract local-aware patch features; (3) Compare each patch with entries in the corresponding memory bank using nearest-neighbor search. The resulting patch-level distances form an anomaly score map, which is upsampled via bilinear interpolation to match the original image resolution. The maximum score in the map is used to determine the image-level anomaly score. This process effectively captures local irregularities while enabling global anomaly detection.

\begin{table}
\caption{Description of four industrial datasets for image anomaly detection.}\label{tab:dataset_description}
\vspace{5pt}
\centering
\scriptsize
\begin{adjustbox}{max width=\textwidth}
\begin{tabular}{@{}clllll@{}}
\toprule
\multicolumn{2}{c|}{\multirow{2}{*}{Class}} &
  \multicolumn{2}{c|}{\# Normal} &
  \multicolumn{1}{c|}{\# Abnormal} &
  \multicolumn{1}{c}{\multirow{2}{*}{\begin{tabular}[c]{@{}c@{}}Anomaly\\      Classes\end{tabular}}} \\ \cmidrule(lr){3-5}
\multicolumn{2}{c|}{} &
  \multicolumn{1}{c}{Train} &
  \multicolumn{1}{c|}{Test} &
  \multicolumn{1}{c|}{Test} &
  \multicolumn{1}{c}{} \\ \midrule
\multicolumn{6}{c}{MVTecAD} \\ \midrule
\multicolumn{1}{c|}{\multirow{10}{*}{\rotatebox{90}{Objects}}} &
  \multicolumn{1}{l|}{Bottle} &
  \multicolumn{1}{c}{209} &
  \multicolumn{1}{c|}{20} &
  \multicolumn{1}{c|}{63} &
  \multicolumn{1}{c}{3} \\
\multicolumn{1}{c|}{} &
  \multicolumn{1}{l|}{Cable} &
  \multicolumn{1}{c}{224} &
  \multicolumn{1}{c|}{58} &
  \multicolumn{1}{c|}{92} &
  \multicolumn{1}{c}{8} \\
\multicolumn{1}{c|}{} &
  \multicolumn{1}{l|}{Capsule} &
  \multicolumn{1}{c}{219} &
  \multicolumn{1}{c|}{23} &
  \multicolumn{1}{c|}{109} &
  \multicolumn{1}{c}{5} \\
\multicolumn{1}{c|}{} &
  \multicolumn{1}{l|}{Hazelnut} &
  \multicolumn{1}{c}{391} &
  \multicolumn{1}{c|}{40} &
  \multicolumn{1}{c|}{70} &
  \multicolumn{1}{c}{4} \\
\multicolumn{1}{c|}{} &
  \multicolumn{1}{l|}{Metal nut} &
  \multicolumn{1}{c}{220} &
  \multicolumn{1}{c|}{22} &
  \multicolumn{1}{c|}{93} &
  \multicolumn{1}{c}{4} \\
\multicolumn{1}{c|}{} &
  \multicolumn{1}{l|}{Pill} &
  \multicolumn{1}{c}{267} &
  \multicolumn{1}{c|}{26} &
  \multicolumn{1}{c|}{141} &
  \multicolumn{1}{c}{7} \\
\multicolumn{1}{c|}{} &
  \multicolumn{1}{l|}{Screw} &
  \multicolumn{1}{c}{320} &
  \multicolumn{1}{c|}{41} &
  \multicolumn{1}{c|}{119} &
  \multicolumn{1}{c}{5} \\
\multicolumn{1}{c|}{} &
  \multicolumn{1}{l|}{Toothbrush} &
  \multicolumn{1}{c}{60} &
  \multicolumn{1}{c|}{12} &
  \multicolumn{1}{c|}{30} &
  \multicolumn{1}{c}{1} \\
\multicolumn{1}{c|}{} &
  \multicolumn{1}{l|}{Transistor} &
  \multicolumn{1}{c}{213} &
  \multicolumn{1}{c|}{60} &
  \multicolumn{1}{c|}{40} &
  \multicolumn{1}{c}{4} \\
\multicolumn{1}{c|}{} &
  \multicolumn{1}{l|}{Zipper} &
  \multicolumn{1}{c}{240} &
  \multicolumn{1}{c|}{32} &
  \multicolumn{1}{c|}{119} &
  \multicolumn{1}{c}{7} \\ \midrule
\multicolumn{1}{c|}{\multirow{5}{*}{\rotatebox{90}{Textures}}} &
  \multicolumn{1}{l|}{Carpet} &
  \multicolumn{1}{c}{280} &
  \multicolumn{1}{c|}{28} &
  \multicolumn{1}{c|}{89} &
  \multicolumn{1}{c}{5} \\
\multicolumn{1}{c|}{} &
  \multicolumn{1}{l|}{Grid} &
  \multicolumn{1}{c}{264} &
  \multicolumn{1}{c|}{21} &
  \multicolumn{1}{c|}{57} &
  \multicolumn{1}{c}{5} \\
\multicolumn{1}{c|}{} &
  \multicolumn{1}{l|}{Leather} &
  \multicolumn{1}{c}{245} &
  \multicolumn{1}{c|}{32} &
  \multicolumn{1}{c|}{92} &
  \multicolumn{1}{c}{5} \\
\multicolumn{1}{c|}{} &
  \multicolumn{1}{l|}{Tile} &
  \multicolumn{1}{c}{230} &
  \multicolumn{1}{c|}{33} &
  \multicolumn{1}{c|}{84} &
  \multicolumn{1}{c}{5} \\
\multicolumn{1}{c|}{} &
  \multicolumn{1}{l|}{Wood} &
  \multicolumn{1}{c}{247} &
  \multicolumn{1}{c|}{19} &
  \multicolumn{1}{c|}{60} &
  \multicolumn{1}{c}{5} \\ \midrule
\multicolumn{2}{c|}{Total} &
  \multicolumn{1}{c}{3,629} &
  \multicolumn{1}{c|}{467} &
  \multicolumn{1}{c|}{1,258} &
  \multicolumn{1}{c}{-} \\ \midrule
\multicolumn{6}{c}{VisA} \\ \midrule
\multicolumn{1}{c|}{\multirow{4}{*}{\rotatebox{90}{\makecell{Complex \\ structure}}}} &
  \multicolumn{1}{l|}{PCB1} &
  \multicolumn{1}{c}{904} &
  \multicolumn{1}{c|}{100} &
  \multicolumn{1}{c|}{100} &
  \multicolumn{1}{c}{4} \\
\multicolumn{1}{c|}{} &
  \multicolumn{1}{l|}{PCB2} &
  \multicolumn{1}{c}{901} &
  \multicolumn{1}{c|}{100} &
  \multicolumn{1}{c|}{100} &
  \multicolumn{1}{c}{4} \\
\multicolumn{1}{c|}{} &
  \multicolumn{1}{l|}{PCB3} &
  \multicolumn{1}{c}{905} &
  \multicolumn{1}{c|}{101} &
  \multicolumn{1}{c|}{100} &
  \multicolumn{1}{c}{4} \\
\multicolumn{1}{c|}{} &
  \multicolumn{1}{l|}{PCB4} &
  \multicolumn{1}{c}{904} &
  \multicolumn{1}{c|}{101} &
  \multicolumn{1}{c|}{100} &
  \multicolumn{1}{c}{7} \\ \midrule
\multicolumn{1}{c|}{\multirow{4}{*}{\rotatebox{90}{\makecell{Multiple \\ instances}}}} &
  \multicolumn{1}{l|}{Macaroni1} &
  \multicolumn{1}{c}{900} &
  \multicolumn{1}{c|}{100} &
  \multicolumn{1}{c|}{100} &
  \multicolumn{1}{c}{5} \\
\multicolumn{1}{c|}{} &
  \multicolumn{1}{l|}{Macaroni2} &
  \multicolumn{1}{c}{900} &
  \multicolumn{1}{c|}{100} &
  \multicolumn{1}{c|}{100} &
  \multicolumn{1}{c}{8} \\
\multicolumn{1}{c|}{} &
  \multicolumn{1}{l|}{Capsules} &
  \multicolumn{1}{c}{542} &
  \multicolumn{1}{c|}{60} &
  \multicolumn{1}{c|}{100} &
  \multicolumn{1}{c}{7} \\
\multicolumn{1}{c|}{} &
  \multicolumn{1}{l|}{Candles} &
  \multicolumn{1}{c}{900} &
  \multicolumn{1}{c|}{100} &
  \multicolumn{1}{c|}{100} &
  \multicolumn{1}{c}{7} \\ \midrule
\multicolumn{1}{c|}{\multirow{4}{*}{\rotatebox{90}{\makecell{Single \\ instance}}}} &
  \multicolumn{1}{l|}{Cashew} &
  \multicolumn{1}{c}{450} &
  \multicolumn{1}{c|}{50} &
  \multicolumn{1}{c|}{100} &
  \multicolumn{1}{c}{9} \\
\multicolumn{1}{c|}{} &
  \multicolumn{1}{l|}{Chewing   gum} &
  \multicolumn{1}{c}{453} &
  \multicolumn{1}{c|}{50} &
  \multicolumn{1}{c|}{100} &
  \multicolumn{1}{c}{6} \\
\multicolumn{1}{c|}{} &
  \multicolumn{1}{l|}{Fryum} &
  \multicolumn{1}{c}{450} &
  \multicolumn{1}{c|}{50} &
  \multicolumn{1}{c|}{100} &
  \multicolumn{1}{c}{8} \\
\multicolumn{1}{c|}{} &
  \multicolumn{1}{l|}{Pipe fryum} &
  \multicolumn{1}{c}{450} &
  \multicolumn{1}{c|}{50} &
  \multicolumn{1}{c|}{100} &
  \multicolumn{1}{c}{6} \\ \midrule
\multicolumn{2}{c|}{Total} &
  \multicolumn{1}{c}{8,659} &
  \multicolumn{1}{c|}{962} &
  \multicolumn{1}{c|}{1,200} &
  \multicolumn{1}{c}{-} \\ \midrule
\multicolumn{6}{c}{MPDD} \\ \midrule
\multicolumn{1}{c|}{\multirow{6}{*}{\rotatebox{90}{Objects}}} &
  \multicolumn{1}{l|}{Bracket   Black} &
  \multicolumn{1}{c}{289} &
  \multicolumn{1}{c|}{32} &
  \multicolumn{1}{c|}{47} &
  \multicolumn{1}{c}{2} \\
\multicolumn{1}{c|}{} &
  \multicolumn{1}{l|}{Bracket Brown} &
  \multicolumn{1}{c}{185} &
  \multicolumn{1}{c|}{26} &
  \multicolumn{1}{c|}{51} &
  \multicolumn{1}{c}{2} \\
\multicolumn{1}{c|}{} &
  \multicolumn{1}{l|}{Bracket White} &
  \multicolumn{1}{c}{110} &
  \multicolumn{1}{c|}{30} &
  \multicolumn{1}{c|}{30} &
  \multicolumn{1}{c}{2} \\
\multicolumn{1}{c|}{} &
  \multicolumn{1}{l|}{Connector} &
  \multicolumn{1}{c}{128} &
  \multicolumn{1}{c|}{30} &
  \multicolumn{1}{c|}{14} &
  \multicolumn{1}{c}{1} \\
\multicolumn{1}{c|}{} &
  \multicolumn{1}{l|}{Metal Plate} &
  \multicolumn{1}{c}{54} &
  \multicolumn{1}{c|}{26} &
  \multicolumn{1}{c|}{71} &
  \multicolumn{1}{c}{3} \\
\multicolumn{1}{c|}{} &
  \multicolumn{1}{l|}{Tubes} &
  \multicolumn{1}{c}{122} &
  \multicolumn{1}{c|}{32} &
  \multicolumn{1}{c|}{69} &
  \multicolumn{1}{c}{1} \\ \midrule
\multicolumn{2}{c|}{Total} &
  \multicolumn{1}{c}{888} &
  \multicolumn{1}{c|}{176} &
  \multicolumn{1}{c|}{282} &
  \multicolumn{1}{c}{-} \\ \midrule
\multicolumn{6}{c}{BTAD} \\ \midrule
\multicolumn{1}{c|}{\multirow{3}{*}{\rotatebox{90}{Objects}}} &
  \multicolumn{1}{l|}{01} &
  \multicolumn{1}{c}{400} &
  \multicolumn{1}{c|}{21} &
  \multicolumn{1}{c|}{49} &
  \multicolumn{1}{c}{1} \\
\multicolumn{1}{c|}{} &
  \multicolumn{1}{l|}{02} &
  \multicolumn{1}{c}{399} &
  \multicolumn{1}{c|}{30} &
  \multicolumn{1}{c|}{200} &
  \multicolumn{1}{c}{1} \\
\multicolumn{1}{c|}{} &
  \multicolumn{1}{l|}{03} &
  \multicolumn{1}{c}{1,000} &
  \multicolumn{1}{c|}{400} &
  \multicolumn{1}{c|}{41} &
  \multicolumn{1}{c}{1} \\ \midrule
\multicolumn{2}{c|}{Total} &
  \multicolumn{1}{c}{1,799} &
  \multicolumn{1}{c|}{451} &
  \multicolumn{1}{c|}{290} &
  \multicolumn{1}{c}{-} \\ \bottomrule
\end{tabular}
\end{adjustbox}
\end{table}

\subsection{Computation Costs for Memory Bank}
\label{sec:complexity}

While PatchCore offers efficient inference by storing only normal data without explicit training, it suffers from high memory and computation costs for large or high-resolution datasets~\cite{ader}. \mbox{HierCore} mitigates this by building class-specific memory banks based on semantic clusters. This reduces both the number of patches and the frequency of feature comparisons. The computational complexity is defined as:

\begin{align}
\Omega(\text{PatchCore}) &= P^2(d + 3r) \\
\Omega(\text{\mbox{HierCore}}) &= \sum_{i=1}^{K} P_i^2(d + 3r) + \Omega(N^2),
\end{align}

\noindent where $P$ is the total number of patches, $d$ is the local feature dimension, $r$ is the sampling ratio for coreset selection, $N$ is the total number of images, and $\Omega(N^2)$ reflects the worst-case complexity of the FINCH algorithm.

Given a stride of 1, the total number of patches is determined by:

\[
P = N \cdot \left( W + 2p - w + 1 \right) \cdot \left( H + 2p - h + 1 \right),
\]

\noindent where $(W, H)$ is the image size, $(w, h)$ is the patch size, $p$ is padding. Since $N \ll P$ in typical datasets, the impact of FINCH clustering is negligible in practice.

Furthermore, \mbox{HierCore} achieves faster inference by using fewer coreset entries per cluster, which reduces the number of comparisons in nearest-neighbor search. This approach makes \mbox{HierCore} more efficient and scalable than PatchCore, particularly in multi-class or large-scale settings.

\section{Experiments}
\label{sec:exp}

\subsection{Experimental Settings}

\paragraph{Image Anomaly Detection Benchmarks}
To evaluate the performance of the proposed \mbox{HierCore} framework under realistic multi-class conditions, we conduct experiments on four widely used industrial benchmark datasets that encompass a diverse range of classes and anomaly types. Specifically, we use MVTec AD~\cite{mvtecad}, VisA~\cite{visa}, MPDD~\cite{mpdd}, and BTAD~\cite{btad}, as summarized in Table~\ref{tab:dataset_description}. These datasets pose challenges for multi-class learning due to significant class imbalance, which makes it difficult for a single model to learn and distinguish class-specific characteristics. Furthermore, all datasets except BTAD include multiple types of anomalies per class, requiring models to generalize from normal data representations to various unseen defect types.

\paragraph{Baselines}
To comprehensively evaluate \mbox{HierCore}, we compare it against both one-class and MC-UIAD baselines. OC-UIAD models typically achieve higher detection performance on a per-class basis, while MC-UIAD models offer efficiency at the expense of performance. This evaluation aims to quantify the extent to which \mbox{HierCore} improves performance within the MC-UIAD setting, narrowing the gap with OC-UIAD approaches.

The OC-UIAD baselines include: DRAEM~\cite{draem}, SimpleNet~\cite{simplenet}, RealNet~\cite{realnet}, RD~\cite{rd}, RD++~\cite{rd++}, DesTSeg~\cite{destseg}, CFLOW-AD~\cite{cflow_ad}, PyramidFlow~\cite{pyramidflow}, CFA~\cite{cfa}, and PatchCore~\cite{patchcore}. The MC-UIAD baselines used for comparison are: UniAD~\cite{uniad}, DiAD~\cite{diad}, ViTAD~\cite{vitad}, InvAD~\cite{invad}, InvAD-lite~\cite{invad}, and MambaAD~\cite{mambaad}.

\paragraph{Implementation Details}
All input images were resized to $256 \times 256$ before training and evaluation. For fair comparison, we followed the training protocol used by Zhang et al.~\cite{ader}, training all models for 100 epochs. In \mbox{HierCore}, we used a Wide-ResNet-50 backbone~\cite{wide_resnet} pre-trained on ImageNet-1k. Semantic features were extracted from the fourth layer of the network, while local features were extracted from the second and third layers. Local aware patch features were extracted using a $3 \times 3$ sliding window with a stride of 1. The coreset for the memory bank was formed by selecting 10\% of all local aware patch features. During inference, the anomaly score for each patch was computed based on the L2 distance to its nearest coreset element in the corresponding memory bank. For efficient nearest neighbor search, we used the GPU-accelerated FAISS library~\cite{faiss}. All experiments were conducted on a system equipped with an Intel i7-8700K CPU, 96GB of RAM, and an NVIDIA GeForce RTX 4090 GPU.

\subsection{Evaluation Protocol}

We conduct a comprehensive set of experiments to evaluate whether the proposed \mbox{HierCore} satisfies the key requirements of MC-UIAD models in practical scenarios. The evaluation is structured in two main directions: (1) performance comparison between MC-UIAD and OC-UIAD models, and (2) validation of whether MC-UIAD models satisfy the proposed requirements under multi-class settings for real-world deployment.

\paragraph{Evaluating Performance of Image Anomaly Detection in Multi-class Settings}
The first set of experiments assesses the performance of \mbox{HierCore} and existing MC-UIAD models under a multi-class setting, comparing them against OC-UIAD baselines. Previous works on MC-UIAD have primarily aimed to minimize performance degradation when transitioning from one-class to multi-class detection. Following this convention, we evaluate detection performance under the following three conditions:

\begin{enumerate}
    \item OC-UIAD models in one-class setting,
    \item OC-UIAD models in multi-class setting,
    \item MC-UIAD models in multi-class setting.
\end{enumerate}

Anomaly detection performance is measured at both the image- and pixel-levels, and results are macro-averaged across all classes. For image-level evaluation, we report three metrics: mean Area Under the Receiver Operating Characteristic Curve (mAUROC), mean Average Precision (mAP)~\cite{draem}, and mean F1-score at the optimal threshold (mF1-max)~\cite{visa}. For pixel-level evaluation, we use five metrics: mAUROC, mAP, mF1-max, mean Area Under the Per-Region-Overlap curve (mAUPRO)~\cite{aupro}, and mean Intersection over Union at the optimal threshold (mIoU-max)~\cite{invad}. In addition, we report an overall score, mean Anomaly Detection score (mAD), defined as the average of all the above metrics.

\paragraph{Evaluating Requirements for Multi-class Image Anomaly Detection}
The second set of experiments aims to verify whether \mbox{HierCore} satisfies the two requirements for real-world MC-UIAD, as defined in Section~\ref{sec:intro}. Specifically, we examine the model's robustness across two representative conditions, based on the availability of class labels during training and evaluation:

\begin{itemize}
    \item \textbf{Condition (1): Unknown → Known \& Unknown.} Training without class labels, evaluation under both settings.
    \item \textbf{Condition (2): Known → Known \& Unknown.} Training with class labels, evaluation under both settings.
\end{itemize}

\begin{table}
\caption{Image- and pixel-level mAD on four industrial datasets. The table presents a comparison between one-class and multi-class unsupervised image anomaly detection models. One-class models are evaluated under both \textcolor{gray}{one-class} and multi-class settings. All experiments are conducted assuming unknown classes during training and known classes during evaluation.}\label{tab:exp_w_oc_mc}
\vspace{5pt}
\centering
\begin{adjustbox}{max width=\textwidth}
\begin{tabular}{@{}cl|cccccccc|cccccccc@{}}
\toprule
\multicolumn{2}{c|}{} &
  \multicolumn{8}{c|}{Image-level mAD} &
  \multicolumn{8}{c}{Pixel-leval mAD} \\ \cmidrule(l){3-18} 
\multicolumn{2}{c|}{\multirow{-2}{*}{Model}} &
  \multicolumn{2}{c|}{MVTecAD} &
  \multicolumn{2}{c|}{VisA} &
  \multicolumn{2}{c|}{MPDD} &
  \multicolumn{2}{c|}{BTAD} &
  \multicolumn{2}{c|}{MVTecAD} &
  \multicolumn{2}{c|}{VisA} &
  \multicolumn{2}{c|}{MPDD} &
  \multicolumn{2}{c}{BTAD} \\ \midrule
\multicolumn{1}{c|}{} &
  DRAEM &
  {\color[HTML]{808080} 0.816} &
  \multicolumn{1}{c|}{0.715} &
  {\color[HTML]{808080} 0.743} &
  \multicolumn{1}{c|}{0.635} &
  {\color[HTML]{808080} 0.778} &
  \multicolumn{1}{c|}{0.538} &
  {\color[HTML]{808080} 0.754} &
  0.759 &
  {\color[HTML]{808080} 0.183} &
  \multicolumn{1}{c|}{0.151} &
  {\color[HTML]{808080} 0.160} &
  \multicolumn{1}{c|}{0.101} &
  {\color[HTML]{808080} 0.154} &
  \multicolumn{1}{c|}{0.141} &
  {\color[HTML]{808080} 0.182} &
  0.158 \\
\multicolumn{1}{c|}{} &
  SimpleNet &
  {\color[HTML]{808080} 0.983} &
  \multicolumn{1}{c|}{0.965} &
  {\color[HTML]{808080} 0.936} &
  \multicolumn{1}{c|}{0.861} &
  {\color[HTML]{808080} 0.950} &
  \multicolumn{1}{c|}{0.894} &
  {\color[HTML]{808080} 0.952} &
  0.946 &
  {\color[HTML]{808080} 0.664} &
  \multicolumn{1}{c|}{0.642} &
  {\color[HTML]{808080} 0.572} &
  \multicolumn{1}{c|}{0.547} &
  {\color[HTML]{808080} 0.524} &
  \multicolumn{1}{c|}{0.553} &
  {\color[HTML]{808080} 0.582} &
  0.561 \\
\multicolumn{1}{c|}{} &
  RealNet &
  {\color[HTML]{808080} 0.970} &
  \multicolumn{1}{c|}{0.899} &
  {\color[HTML]{808080} 0.917} &
  \multicolumn{1}{c|}{0.752} &
  {\color[HTML]{808080} 0.792} &
  \multicolumn{1}{c|}{0.879} &
  {\color[HTML]{808080} 0.949} &
  0.926 &
  {\color[HTML]{808080} 0.692} &
  \multicolumn{1}{c|}{0.496} &
  {\color[HTML]{808080} 0.540} &
  \multicolumn{1}{c|}{0.300} &
  {\color[HTML]{808080} 0.458} &
  \multicolumn{1}{c|}{0.511} &
  {\color[HTML]{808080} 0.643} &
  0.550 \\
\multicolumn{1}{c|}{} &
  RD &
  {\color[HTML]{808080} 0.985} &
  \multicolumn{1}{c|}{0.955} &
  {\color[HTML]{808080} 0.949} &
  \multicolumn{1}{c|}{0.903} &
  {\color[HTML]{808080} 0.944} &
  \multicolumn{1}{c|}{0.922} &
  {\color[HTML]{808080} 0.941} &
  0.950 &
  {\color[HTML]{808080} 0.688} &
  \multicolumn{1}{c|}{0.652} &
  {\color[HTML]{808080} 0.618} &
  \multicolumn{1}{c|}{0.591} &
  {\color[HTML]{808080} 0.630} &
  \multicolumn{1}{c|}{0.615} &
  {\color[HTML]{808080} 0.672} &
  {\ul 0.679} \\
\multicolumn{1}{c|}{} &
  RD++ &
  {\color[HTML]{808080} 0.987} &
  \multicolumn{1}{c|}{0.977} &
  {\color[HTML]{808080} 0.943} &
  \multicolumn{1}{c|}{0.929} &
  {\color[HTML]{808080} 0.921} &
  \multicolumn{1}{c|}{0.913} &
  {\color[HTML]{808080} 0.951} &
  0.955 &
  {\color[HTML]{808080} 0.695} &
  \multicolumn{1}{c|}{0.690} &
  {\color[HTML]{808080} 0.616} &
  \multicolumn{1}{c|}{0.620} &
  {\color[HTML]{808080} 0.614} &
  \multicolumn{1}{c|}{0.629} &
  {\color[HTML]{808080} 0.672} &
  0.678 \\
\multicolumn{1}{c|}{} &
  DesTSeg &
  {\color[HTML]{808080} 0.970} &
  \multicolumn{1}{c|}{0.971} &
  {\color[HTML]{808080} 0.904} &
  \multicolumn{1}{c|}{0.893} &
  {\color[HTML]{808080} 0.921} &
  \multicolumn{1}{c|}{0.908} &
  {\color[HTML]{808080} 0.915} &
  0.937 &
  {\color[HTML]{808080} 0.731} &
  \multicolumn{1}{c|}{0.735} &
  {\color[HTML]{808080} 0.534} &
  \multicolumn{1}{c|}{0.549} &
  {\color[HTML]{808080} 0.529} &
  \multicolumn{1}{c|}{0.476} &
  {\color[HTML]{808080} 0.547} &
  0.541 \\
\multicolumn{1}{c|}{} &
  CFLOW-AD &
  {\color[HTML]{808080} 0.961} &
  \multicolumn{1}{c|}{0.939} &
  {\color[HTML]{808080} 0.892} &
  \multicolumn{1}{c|}{0.867} &
  {\color[HTML]{808080} 0.865} &
  \multicolumn{1}{c|}{0.792} &
  {\color[HTML]{808080} 0.947} &
  0.914 &
  {\color[HTML]{808080} 0.670} &
  \multicolumn{1}{c|}{0.623} &
  {\color[HTML]{808080} 0.567} &
  \multicolumn{1}{c|}{0.561} &
  {\color[HTML]{808080} 0.564} &
  \multicolumn{1}{c|}{0.521} &
  {\color[HTML]{808080} 0.581} &
  0.598 \\
\multicolumn{1}{c|}{} &
  PyramidFlow &
  {\color[HTML]{808080} 0.909} &
  \multicolumn{1}{c|}{0.804} &
  {\color[HTML]{808080} 0.892} &
  \multicolumn{1}{c|}{0.663} &
  {\color[HTML]{808080} 0.850} &
  \multicolumn{1}{c|}{0.767} &
  {\color[HTML]{808080} 0.797} &
  0.837 &
  {\color[HTML]{808080} 0.605} &
  \multicolumn{1}{c|}{0.369} &
  {\color[HTML]{808080} 0.501} &
  \multicolumn{1}{c|}{0.284} &
  {\color[HTML]{808080} 0.535} &
  \multicolumn{1}{c|}{0.441} &
  {\color[HTML]{808080} 0.523} &
  0.460 \\
\multicolumn{1}{c|}{} &
  CFA &
  {\color[HTML]{808080} 0.946} &
  \multicolumn{1}{c|}{0.735} &
  {\color[HTML]{808080} 0.868} &
  \multicolumn{1}{c|}{0.716} &
  {\color[HTML]{808080} 0.872} &
  \multicolumn{1}{c|}{0.850} &
  {\color[HTML]{808080} 0.955} &
  0.946 &
  {\color[HTML]{808080} 0.624} &
  \multicolumn{1}{c|}{0.231} &
  {\color[HTML]{808080} 0.516} &
  \multicolumn{1}{c|}{0.395} &
  {\color[HTML]{808080} 0.441} &
  \multicolumn{1}{c|}{0.395} &
  {\color[HTML]{808080} 0.648} &
  0.594 \\
\multicolumn{1}{c|}{\multirow{-10}{*}{\rotatebox{90}{One-class}}} &
  PatchCore &
  {\color[HTML]{808080} 0.992} &
  \multicolumn{1}{c|}{\textbf{0.992}} &
  {\color[HTML]{808080} 0.953} &
  \multicolumn{1}{c|}{\textbf{0.953}} &
  {\color[HTML]{808080} 0.949} &
  \multicolumn{1}{c|}{\textbf{0.952}} &
  {\color[HTML]{808080} 0.952} &
  0.956 &
  {\color[HTML]{808080} 0.751} &
  \multicolumn{1}{c|}{{\ul 0.745}} &
  {\color[HTML]{808080} 0.653} &
  \multicolumn{1}{c|}{\textbf{0.656}} &
  {\color[HTML]{808080} 0.656} &
  \multicolumn{1}{c|}{\textbf{0.659}} &
  {\color[HTML]{808080} 0.652} &
  0.653 \\ \midrule
\multicolumn{1}{c|}{} &
  UniAD &
  - &
  \multicolumn{1}{c|}{0.950} &
  - &
  \multicolumn{1}{c|}{0.884} &
  - &
  \multicolumn{1}{c|}{0.752} &
  - &
  {\ul 0.960} &
  - &
  \multicolumn{1}{c|}{0.616} &
  - &
  \multicolumn{1}{c|}{0.564} &
  - &
  \multicolumn{1}{c|}{0.442} &
  - &
  0.631 \\
\multicolumn{1}{c|}{} &
  DiAD &
  - &
  \multicolumn{1}{c|}{0.927} &
  - &
  \multicolumn{1}{c|}{0.867} &
  - &
  \multicolumn{1}{c|}{0.754} &
  - &
  0.904 &
  - &
  \multicolumn{1}{c|}{0.468} &
  - &
  \multicolumn{1}{c|}{0.366} &
  - &
  \multicolumn{1}{c|}{0.377} &
  - &
  0.448 \\
\multicolumn{1}{c|}{} &
  ViTAD &
  - &
  \multicolumn{1}{c|}{0.980} &
  - &
  \multicolumn{1}{c|}{0.895} &
  - &
  \multicolumn{1}{c|}{0.881} &
  - &
  0.950 &
  - &
  \multicolumn{1}{c|}{0.691} &
  - &
  \multicolumn{1}{c|}{0.577} &
  - &
  \multicolumn{1}{c|}{0.582} &
  - &
  0.665 \\
\multicolumn{1}{c|}{} &
  InvAD &
  - &
  \multicolumn{1}{c|}{{\ul 0.983}} &
  - &
  \multicolumn{1}{c|}{{\ul 0.944}} &
  - &
  \multicolumn{1}{c|}{0.934} &
  - &
  \textbf{0.961} &
  - &
  \multicolumn{1}{c|}{0.700} &
  - &
  \multicolumn{1}{c|}{0.628} &
  - &
  \multicolumn{1}{c|}{0.632} &
  - &
  \textbf{0.691} \\
\multicolumn{1}{c|}{} &
  InvAD-lite &
  - &
  \multicolumn{1}{c|}{0.980} &
  - &
  \multicolumn{1}{c|}{0.936} &
  - &
  \multicolumn{1}{c|}{0.911} &
  - &
  0.950 &
  - &
  \multicolumn{1}{c|}{0.690} &
  - &
  \multicolumn{1}{c|}{0.611} &
  - &
  \multicolumn{1}{c|}{0.610} &
  - &
  0.676 \\
\multicolumn{1}{c|}{} &
  MambaAD &
  - &
  \multicolumn{1}{c|}{0.978} &
  - &
  \multicolumn{1}{c|}{0.927} &
  - &
  \multicolumn{1}{c|}{0.903} &
  - &
  0.942 &
  - &
  \multicolumn{1}{c|}{0.687} &
  - &
  \multicolumn{1}{c|}{0.605} &
  - &
  \multicolumn{1}{c|}{0.585} &
  - &
  0.643 \\
\multicolumn{1}{c|}{\multirow{-7}{*}{\rotatebox{90}{Multi-class}}} &
  HierCore   (Ours) &
  - &
  \multicolumn{1}{c|}{\textbf{0.992}} &
  - &
  \multicolumn{1}{c|}{\textbf{0.953}} &
  - &
  \multicolumn{1}{c|}{{\ul 0.948}} &
  - &
  0.952 &
  - &
  \multicolumn{1}{c|}{\textbf{0.748}} &
  - &
  \multicolumn{1}{c|}{{\ul 0.649}} &
  - &
  \multicolumn{1}{c|}{{\ul 0.655}} &
  - &
  0.652 \\ \bottomrule
\end{tabular}
\end{adjustbox}
\end{table}

In Condition (2), which requires models trained with class labels, we exclude MINT-AD~\cite{mint_ad} due to the lack of a publicly available implementation. Other existing MC-UIAD methods are structurally label-agnostic and do not support training with class labels. As a result, only \mbox{HierCore} is evaluated under this condition.

Since the presence of class labels affects how the decision threshold for anomaly detection is defined, performance may vary between conditions. In this study, we define the optimal threshold $\hat{T}$ as:

\[
\hat{T} = \arg\max_T \text{F1-score}(T), \quad \text{based on the precision-recall curve}.
\]

Using this criterion, we compare detection performance under each condition to validate whether \mbox{HierCore} maintains stable results regardless of label availability.

\subsection{Evaluation of Performance Gap between One-class and Multi-class Image Anomaly Detection}

Table~\ref{tab:exp_w_oc_mc} presents the performance comparison between OC-UIAD and MC-UIAD models on four industrial datasets: MVTec AD, VisA, MPDD, and BTAD. All evaluations were conducted under the known-class setting, where class information is available during testing.

As expected, OC-UIAD models, which are inherently designed for one-class settings, generally experience performance degradation when extended to multi-class scenarios. However, PatchCore is a notable exception: despite being an OC-UIAD model, it demonstrates stable and robust performance across both one-class and multi-class settings. This result can be attributed to PatchCore’s memory bank-based architecture, which does not rely on end-to-end training and is not affected by the \textit{identical shortcut} problem that commonly affects reconstruction-based models in multi-class contexts.

While MC-UIAD models tend to outperform OC-UIAD models in multi-class settings, this is not universally the case. For instance, reconstruction-based OC-UIAD models such as RD and RD++ achieve competitive performance compared to several MC-UIAD baselines. PatchCore, although not designed for multi-class settings, achieves top-tier performance on MVTec AD, VisA, and MPDD at both image- and pixel-level evaluations. Its only relative weakness is observed on the BTAD dataset.

The proposed \mbox{HierCore} framework shows a similar trend to PatchCore. Across most datasets (MVTec AD, VisA, MPDD), \mbox{HierCore} outperforms existing MC-UIAD models in both image- and pixel-level anomaly detection. Its performance is particularly strong when class labels are not used during training, further confirming its robustness. On BTAD, while the performance is slightly lower than on other datasets, \mbox{HierCore} remains competitive.

Detailed class-wise performance metrics for all models across the four datasets are provided in \ref{appendix:oc_mc}.

\subsection{Evaluation of Requirements for Multi-Class Unsupervised Image Anomaly Detection}

\subsubsection{Unknown to Known and Unknown}

\begin{table}
\caption{Image- and pixel-level anomaly detection performance on four industrial datasets based on average F1-score with optimal thresholding. Models are trained under an unknown classes setting. During evaluation, F1-scores for known classes are computed using class-specific optimal thresholds, while a single optimal threshold is used for unknown-class evaluation. Diff. Ratio indicates the ratio of F1-scores under the unknown-class evaluation to those under the known-class evaluation.}\label{tab:exp_u_to_uk}
\vspace{5pt}
\centering
\begin{adjustbox}{max width=\textwidth}
\begin{tabular}{@{}c|c|cccccc|cccccc@{}}
\toprule
\multirow{2}{*}{Dataset} &
  \multirow{2}{*}{Evaluation} &
  \multicolumn{6}{c|}{Image-level} &
  \multicolumn{6}{c}{Pixel-level} \\ \cmidrule(l){3-14} 
 &
   &
  UniAD &
  ViTAD &
  InvAD &
  MambaAD &
  PatchCore &
  HierCore &
  UniAD &
  ViTAD &
  InvAD &
  MambaAD &
  PatchCore &
  HierCore \\ \midrule
\multirow{3}{*}{MVTecAD} &
  Known &
  91.9 &
  96.0 &
  96.9 &
  96.2 &
  {\ul 97.8} &
  \textbf{97.9} &
  48.3 &
  58.9 &
  59.3 &
  57.6 &
  {\ul 65.4} &
  \textbf{65.8} \\
 &
  Unknown &
  87.3 &
  89.2 &
  90.0 &
  88.4 &
  {\ul 91.7} &
  \textbf{97.9} &
  43.7 &
  49.0 &
  51.8 &
  51.5 &
  {\ul 58.7} &
  \textbf{65.9} \\ \cmidrule(l){2-14} 
 &
  Diff.   Ratio &
  {\ul 95.1\%} &
  92.9\% &
  92.9\% &
  91.9\% &
  93.8\% &
  \textbf{100.0\%} &
  {\ul 90.4\%} &
  83.2\% &
  87.2\% &
  89.4\% &
  89.8\% &
  \textbf{100.1\%} \\ \midrule
\multirow{3}{*}{VisA} &
  Known &
  84.6 &
  85.9 &
  91.5 &
  88.7 &
  \textbf{92.5} &
  {\ul 92.2} &
  39.3 &
  41.3 &
  47.2 &
  43.8 &
  \textbf{52.0} &
  {\ul 50.9} \\
 &
  Unknown &
  81.5 &
  81.0 &
  86.3 &
  85.5 &
  {\ul 90.0} &
  \textbf{92.2} &
  35.3 &
  34.4 &
  42.3 &
  39.4 &
  {\ul 45.6} &
  \textbf{50.9} \\ \cmidrule(l){2-14} 
 &
  Diff.   Ratio &
  96.3\% &
  94.2\% &
  94.3\% &
  96.4\% &
  {\ul 97.2\%} &
  \textbf{100.0\%} &
  {\ul 90.0\%} &
  83.3\% &
  89.7\% &
  90.0\% &
  87.8\% &
  \textbf{100.0\%} \\ \midrule
\multirow{3}{*}{MPDD} &
  Known &
  78.8 &
  84.9 &
  84.9 &
  91.2 &
  \textbf{92.2} &
  {\ul 91.8} &
  20.5 &
  37.7 &
  37.7 &
  46.2 &
  \textbf{50.6} &
  {\ul 50.3} \\
 &
  Unknown &
  76.9 &
  77.0 &
  77.0 &
  {\ul 84.0} &
  83.8 &
  \textbf{91.7} &
  15.4 &
  22.5 &
  22.5 &
  {\ul 40.5} &
  28.6 &
  \textbf{50.5} \\ \cmidrule(l){2-14} 
 &
  Diff.   Ratio &
  {\ul 97.5\%} &
  90.8\% &
  90.8\% &
  92.1\% &
  90.9\% &
  \textbf{99.9\%} &
  75.4\% &
  59.6\% &
  59.6\% &
  {\ul 87.5\%} &
  56.5\% &
  \textbf{100.5\%} \\ \midrule
\multirow{3}{*}{BTAD} &
  Known &
  93.1 &
  92.9 &
  \textbf{93.9} &
  92.7 &
  {\ul 93.7} &
  92.8 &
  54.1 &
  {\ul 56.9} &
  \textbf{61.4} &
  55.7 &
  56.5 &
  56.5 \\
 &
  Unknown &
  {\ul 81.0} &
  77.9 &
  80.3 &
  75.7 &
  78.4 &
  \textbf{92.8} &
  53.0 &
  22.8 &
  53.3 &
  49.3 &
  {\ul 55.2} &
  \textbf{56.5} \\ \cmidrule(l){2-14} 
 &
  Diff.   Ratio &
  {\ul 87.0\%} &
  83.9\% &
  85.5\% &
  81.7\% &
  83.7\% &
  \textbf{100.0\%} &
  {\ul 98.0\%} &
  40.0\% &
  86.8\% &
  88.6\% &
  97.7\% &
  \textbf{100.0\%} \\ \bottomrule
\end{tabular}
\end{adjustbox}
\end{table}

Table~\ref{tab:exp_u_to_uk} shows the performance variation of models trained without class labels under two different evaluation conditions: with and without class label availability. This experiment is designed to assess the robustness of anomaly detection performance in realistic scenarios where class labels may not be available at inference time.

Among the MC-UIAD models, UniAD exhibits a relatively small performance drop under the unknown evaluation condition. However, its overall detection accuracy remains consistently lower than other models, even when class labels are provided at evaluation. This suggests that UniAD’s apparent stability may be due to low baseline performance, rather than true robustness to label absence.

PatchCore achieves performance comparable to \mbox{HierCore} when class labels are available during evaluation. However, it shows a significant performance drop under the unknown evaluation condition. This indicates a strong dependency on class-specific thresholding, which limits its applicability in real-world settings where such labels are not accessible.

In contrast, \mbox{HierCore} maintains high and consistent anomaly detection performance regardless of whether class labels are available during evaluation. This result highlights the reliability of \mbox{HierCore} in scenarios where class information is unavailable, an important property for real-world MC-UIAD applications.

Detailed class-wise performance comparisons for all four datasets are provided in \ref{appendix:u_to_uk}.

\subsubsection{Known to Known and Unknown}

\begin{table}
\caption{Average F1-score on the optimal threshold. Known training setting. Optimal threshold per class is applied in Known evaluation. Optimal threshold for all samples is applied in Unknown evaluation. Diff. Ratio is the ratio of F1-score in Unknown with repect to Known.}\label{tab:exp_k_to_uk}
\vspace{5pt}
\centering
\footnotesize
\begin{adjustbox}{max width=\textwidth}
\begin{tabular}{@{}c|c|cc@{}}
\toprule
\multirow{2}{*}{Dataset} & \multirow{2}{*}{Evaluation} & \multicolumn{2}{c}{HierCore} \\
                         &                             & Image-level   & Pixel-level  \\ \midrule
\multirow{3}{*}{MVTecAD} & Known                       & 97.9          & 66.1         \\
                         & Unknown                     & 97.9          & 65.9         \\ \cmidrule(l){2-4} 
                         & Diff.   Ratio               & 100.0\%       & 99.6\%       \\ \midrule
\multirow{3}{*}{VisA}    & Known                       & 92.2          & 51.3         \\
                         & Unknown                     & 92.2          & 50.9         \\ \cmidrule(l){2-4} 
                         & Diff.   Ratio               & 100.0\%       & 99.2\%       \\ \midrule
\multirow{3}{*}{MPDD}    & Known                       & 91.8          & 50.4         \\
                         & Unknown                     & 91.8          & 48.0         \\ \cmidrule(l){2-4} 
                         & Diff.   Ratio               & 100.0\%       & 95.2\%       \\ \midrule
\multirow{3}{*}{BTAD}    & Known                       & 92.8          & 56.5         \\
                         & Unknown                     & 92.8          & 56.5         \\ \cmidrule(l){2-4} 
                         & Diff.   Ratio               & 100.0\%       & 100.0\%      \\ \bottomrule
\end{tabular}
\end{adjustbox}
\end{table}

When class labels are available during training, \mbox{HierCore} constructs a separate memory bank for each class. During evaluation, if class labels are also available, each input is directly matched to its corresponding class-specific memory bank, and anomaly detection is performed accordingly. This is functionally similar to deploying a separate PatchCore model for each class.

In contrast, under the unknown evaluation setting, where class labels are not accessible, \mbox{HierCore} estimates the most likely class cluster for a given input using its semantic embedding and the key vector computed as described in Eq.~(2). Anomaly detection is then carried out using the corresponding estimated cluster's memory bank.

Table~\ref{tab:exp_k_to_uk} presents the results of this setting, in which \mbox{HierCore} is trained with known class labels and evaluated both with and without class information. While performance robustness slightly decreases compared to the results in Table~\ref{tab:exp_u_to_uk}, \mbox{HierCore} still outperforms other models' class-aware performance even when evaluated without class labels.

These results suggest that having class labels during training can enhance detection performance by enabling better modeling of class-specific characteristics. However, when comparing Table~\ref{tab:exp_u_to_uk} and Table~\ref{tab:exp_k_to_uk}, it becomes evident that the semantically guided memory bank construction used in label-agnostic training leads to more stable and reliable performance across both evaluation conditions. This highlights the practical advantage of \mbox{HierCore}’s semantic clustering approach for handling real-world scenarios where class information may be missing or unreliable.

\subsection{Visualization of Semantic Centroids and Test Samples}

\begin{figure*}[t!]
\begin{center}
\includegraphics[width=1\linewidth]{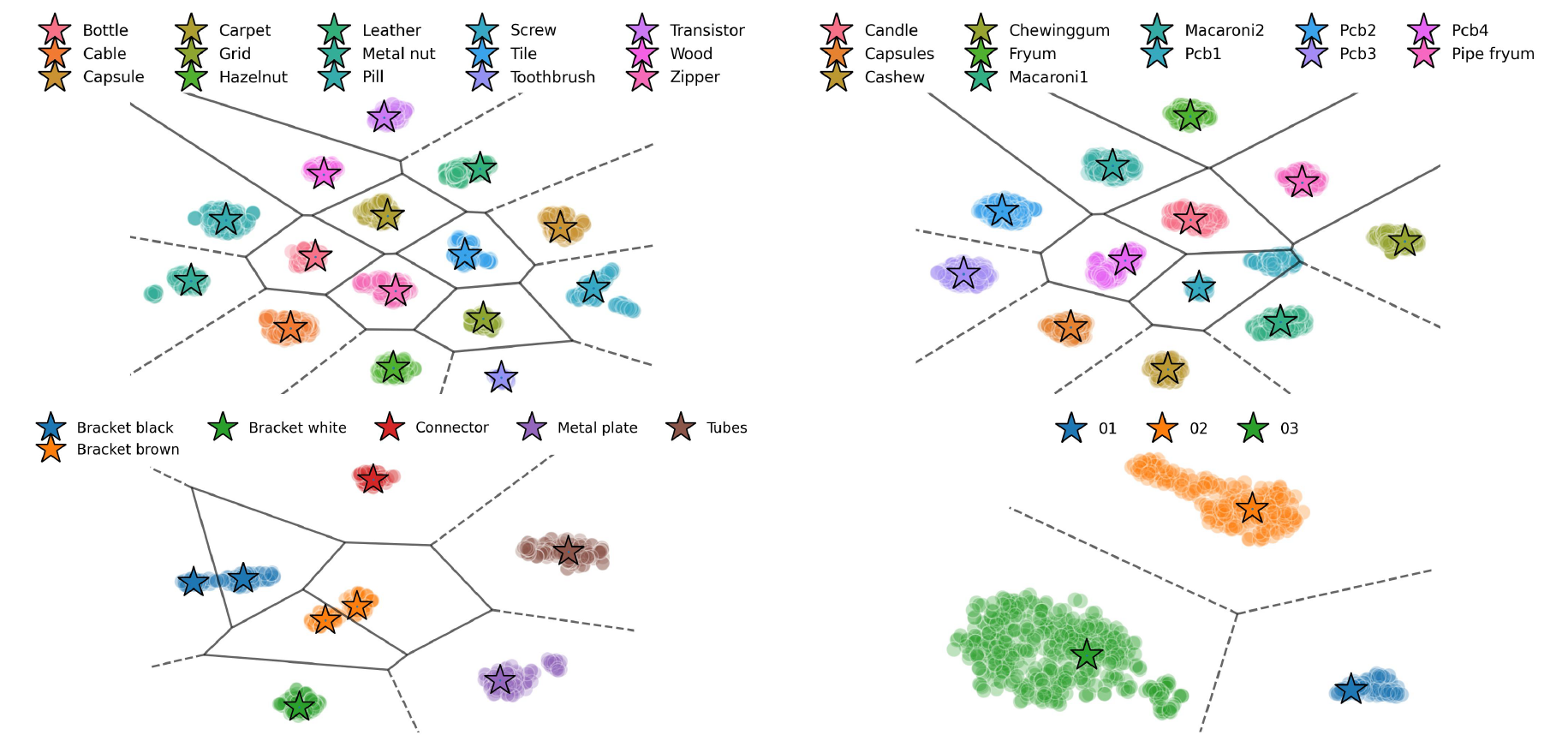}
\end{center}
\vspace{-20pt}
\caption{Visualization of semantic cluster and test samples. Unknown training setting.  The star marks are the centroid of semantic clusters. The circles are test samples. The colors mean class for each cluster and sample.}
\label{fig:clusters}
\end{figure*}

In this section, we verify the effectiveness of \mbox{HierCore}’s approach of separating normal data based on semantic features and organizing memory banks accordingly. Figure~\ref{fig:clusters} provides a 2D visualization of the clustering results obtained from FINCH, using semantic embeddings of normal samples. We apply t-SNE for dimensionality reduction to reveal the structural grouping of the data.

The visualization reveals an interesting phenomenon: in the MPDD dataset, for instance, normal samples that share the same class label are often distributed across multiple distinct semantic clusters. This indicates that even within a single class, normal data can be semantically diverse and may behave as different classes in the embedding space. Such diversity suggests that semantic representations may offer a more meaningful structure for anomaly detection than rigid class labels.

This observation aligns with our empirical findings. For example, in Table~\ref{tab:exp_u_to_uk}, \mbox{HierCore} occasionally achieves better performance under unknown evaluation conditions (i.e., without class labels) than under known ones. This supports the idea that semantic clustering may lead to more effective anomaly detection than label-based grouping. In contrast, Table~\ref{tab:exp_k_to_uk} shows a slight performance drop when models trained with class labels are evaluated without them. This can be attributed to a misalignment between semantic clusters and class labels, suggesting that class labels may not always reflect semantic structure.

These findings collectively demonstrate that \mbox{HierCore}’s memory bank construction based on semantic clustering provides a more robust and generalizable foundation for anomaly detection than approaches strictly based on predefined class labels.

\subsection{Memory Bank and Inference Time}

\begin{table}
\caption{Comparison computation costs for PatchCore and HierCore. The computation time of HierCore for Memory Bank is the sum of the computation times of Memory Bank and FINCH.}\label{tab:exp_computation}
\vspace{5pt}
\centering
\begin{adjustbox}{max width=\textwidth}
\begin{tabular}{@{}l|cc|cc|cc@{}}
\toprule
\multirow{2}{*}{Dataset} & \multicolumn{2}{c|}{Memory Bank (minutes)} & \multicolumn{2}{c|}{Evaluation (seconds)} & \multicolumn{2}{c}{Inference (FPS)} \\ \cmidrule(l){2-7} 
        & PatchCore & HierCore & PatchCore & HierCore & PatchCore & HierCore \\ \midrule
MVTecAD & 317       & 24 + 1   & 196       & 96       & 8.82      & 17.98    \\
VisA    & 1,886     & 154 + 2  & 462       & 131      & 4.67      & 16.51    \\
MPDD    & 17        & 3 + 0    & 29        & 25       & 15.89     & 18.09    \\
BTAD    & 75        & 30 + 0   & 58        & 46       & 12.82     & 15.97    \\ \bottomrule
\end{tabular}
\end{adjustbox}
\end{table}

\mbox{HierCore} constructs memory banks by clustering input data based on semantic similarity and assigning a separate memory bank to each cluster. In contrast, PatchCore builds a single global memory bank using all normal data, regardless of class or semantic grouping. As discussed in Section~\ref{sec:complexity}, this structural difference allows \mbox{HierCore} to significantly reduce the computational cost of memory bank construction and achieve more efficient nearest-neighbor search, enabling faster anomaly inference.

To quantify this efficiency, Table~\ref{tab:exp_computation} compares three metrics across four industrial datasets: memory bank construction time (in hours), total inference time for the entire evaluation dataset (in seconds), and the number of frames processed per second (FPS) during inference.

On datasets with a large number of classes, such as MVTec AD, \mbox{HierCore} achieves approximately 13$\times$ faster memory bank construction compared to PatchCore. For datasets with a large volume of data, such as VisA, the difference is even more pronounced: while PatchCore takes nearly 30 hours to construct the memory bank, \mbox{HierCore} completes the process over 12 times faster. In terms of inference time, \mbox{HierCore} records an average of 3.5$\times$ speed-up compared to PatchCore. Similarly, \mbox{HierCore} shows a 3.5$\times$ improvement in FPS on average. Notably, PatchCore exhibits significant FPS variance across datasets, with up to a 3.4$\times$ difference between its fastest and slowest cases. In contrast, \mbox{HierCore} maintains stable performance across datasets, with a maximum FPS variation within 1.1$\times$.

These results highlight that \mbox{HierCore} is not only faster but also more consistent in runtime efficiency, making it highly suitable for real-time anomaly detection in multi-class or large-scale industrial settings.

\section{Conclusion}
\label{sec:conclusion}

This study addressed a practical challenge in extending OC-UIAD methods to multi-class scenarios, going beyond mere performance improvement to focus on flexible applicability depending on the availability of class information during training and evaluation. To this end, we defined two requirements for multi-class image anomaly detection and re-evaluated existing MC-UIAD approaches in terms of their ability to meet these criteria.

We proposed \mbox{HierCore}, a novel memory-based framework that can operate effectively regardless of whether class labels are available. Extensive experiments on four industrial benchmark datasets demonstrated that \mbox{HierCore} achieves robust and consistent performance across both known and unknown training and evaluation settings. While most previous MC-UIAD methods rely on reconstruction-based approaches, our study explored the scalability and robustness of memory bank-based methods. Notably, we found that PatchCore, although originally designed as an OC-UIAD method, retains strong performance even in multi-class environments.

\mbox{HierCore} addresses the main limitations of PatchCore by clustering normal data based on semantic features and constructing independent memory banks for each cluster. This hierarchical structure significantly reduces the computational burden and inference latency caused by scaling to large datasets. Our experiments showed that even when using only 10\% of patch features to build the memory bank, \mbox{HierCore} maintained high anomaly detection accuracy.

We hope that this work encourages further research into MC-UIAD methods that satisfy both practical requirements proposed herein and are adaptable to diverse real-world industrial settings.

\bibliography{egbib}

\appendix

\section{Evaluation of Performance Gap between One-class and Multi-class Image Anomaly Detection}
\label{appendix:oc_mc}

\begin{table}[H]
\caption{Anomaly detection performance on the MVTecAD dataset. The table presents a comparison between one-class and multi-class unsupervised image anomaly detection models. One-class models are evaluated under both \textcolor{gray}{one-class} and multi-class settings. All experiments are conducted assuming unknown classes during training and known classes during evaluation.}\label{tab:exp_w_oc_mc_mvtecad}
\vspace{5pt}
\centering
\begin{adjustbox}{max width=\textwidth}
\begin{tabular}{@{}cl|lccccccc|cccccccccccc@{}}
\toprule
\multicolumn{2}{c|}{} &
  \multicolumn{8}{c|}{Image-level} &
  \multicolumn{12}{c}{Pixel-level} \\ \cmidrule(l){3-22} 
\multicolumn{2}{c|}{\multirow{-2}{*}{Model}} &
  \multicolumn{2}{c|}{mAUROC} &
  \multicolumn{2}{c|}{mAP} &
  \multicolumn{2}{c|}{mF1-max} &
  \multicolumn{2}{c|}{mAD} &
  \multicolumn{2}{c|}{mAUROC} &
  \multicolumn{2}{c|}{mAP} &
  \multicolumn{2}{c|}{mF1-max} &
  \multicolumn{2}{c|}{mAUPRO} &
  \multicolumn{2}{c|}{mIoU-max} &
  \multicolumn{2}{c}{mAD} \\ \midrule
\multicolumn{1}{c|}{} &
  DRAEM &
  {\color[HTML]{808080} 0.716} &
  \multicolumn{1}{c|}{0.545} &
  {\color[HTML]{808080} 0.864} &
  \multicolumn{1}{c|}{0.763} &
  {\color[HTML]{808080} 0.868} &
  \multicolumn{1}{c|}{0.836} &
  {\color[HTML]{808080} 0.816} &
  0.715 &
  {\color[HTML]{808080} 0.567} &
  \multicolumn{1}{c|}{0.476} &
  {\color[HTML]{808080} 0.083} &
  \multicolumn{1}{c|}{0.032} &
  {\color[HTML]{808080} 0.114} &
  \multicolumn{1}{c|}{0.067} &
  {\color[HTML]{808080} 0.254} &
  \multicolumn{1}{c|}{0.143} &
  {\color[HTML]{808080} 0.061} &
  \multicolumn{1}{c|}{0.035} &
  {\color[HTML]{808080} 0.216} &
  0.151 \\
\multicolumn{1}{c|}{} &
  SimpleNet &
  {\color[HTML]{808080} 0.981} &
  \multicolumn{1}{c|}{0.954} &
  \multicolumn{1}{l}{{\color[HTML]{808080} 0.993}} &
  \multicolumn{1}{c|}{0.983} &
  \multicolumn{1}{l}{{\color[HTML]{808080} 0.976}} &
  \multicolumn{1}{c|}{0.957} &
  {\color[HTML]{808080} 0.983} &
  0.965 &
  \multicolumn{1}{l}{{\color[HTML]{808080} 0.973}} &
  \multicolumn{1}{c|}{0.968} &
  \multicolumn{1}{l}{{\color[HTML]{808080} 0.525}} &
  \multicolumn{1}{c|}{0.488} &
  \multicolumn{1}{l}{{\color[HTML]{808080} 0.550}} &
  \multicolumn{1}{c|}{0.519} &
  \multicolumn{1}{l}{{\color[HTML]{808080} 0.882}} &
  \multicolumn{1}{c|}{0.869} &
  \multicolumn{1}{l}{{\color[HTML]{808080} 0.392}} &
  \multicolumn{1}{c|}{0.364} &
  {\color[HTML]{808080} 0.664} &
  0.642 \\
\multicolumn{1}{c|}{} &
  RealNet &
  {\color[HTML]{808080} 0.966} &
  \multicolumn{1}{c|}{0.848} &
  {\color[HTML]{808080} 0.986} &
  \multicolumn{1}{c|}{0.941} &
  {\color[HTML]{808080} 0.959} &
  \multicolumn{1}{c|}{0.909} &
  {\color[HTML]{808080} 0.970} &
  0.899 &
  {\color[HTML]{808080} 0.932} &
  \multicolumn{1}{c|}{0.726} &
  {\color[HTML]{808080} 0.601} &
  \multicolumn{1}{c|}{0.482} &
  {\color[HTML]{808080} 0.612} &
  \multicolumn{1}{c|}{0.414} &
  {\color[HTML]{808080} 0.867} &
  \multicolumn{1}{c|}{0.568} &
  {\color[HTML]{808080} 0.449} &
  \multicolumn{1}{c|}{0.288} &
  {\color[HTML]{808080} 0.692} &
  0.496 \\
\multicolumn{1}{c|}{} &
  RD &
  {\color[HTML]{808080} 0.985} &
  \multicolumn{1}{c|}{0.936} &
  {\color[HTML]{808080} 0.993} &
  \multicolumn{1}{c|}{0.972} &
  {\color[HTML]{808080} 0.976} &
  \multicolumn{1}{c|}{0.956} &
  {\color[HTML]{808080} 0.985} &
  0.955 &
  {\color[HTML]{808080} 0.974} &
  \multicolumn{1}{c|}{0.958} &
  {\color[HTML]{808080} 0.540} &
  \multicolumn{1}{c|}{0.482} &
  {\color[HTML]{808080} 0.578} &
  \multicolumn{1}{c|}{0.536} &
  {\color[HTML]{808080} 0.934} &
  \multicolumn{1}{c|}{0.912} &
  {\color[HTML]{808080} 0.414} &
  \multicolumn{1}{c|}{0.370} &
  {\color[HTML]{808080} 0.688} &
  0.652 \\
\multicolumn{1}{c|}{} &
  RD++ &
  {\color[HTML]{808080} 0.986} &
  \multicolumn{1}{c|}{0.979} &
  {\color[HTML]{808080} 0.994} &
  \multicolumn{1}{c|}{0.988} &
  {\color[HTML]{808080} 0.980} &
  \multicolumn{1}{c|}{0.964} &
  {\color[HTML]{808080} 0.987} &
  0.977 &
  {\color[HTML]{808080} 0.977} &
  \multicolumn{1}{c|}{0.973} &
  {\color[HTML]{808080} 0.555} &
  \multicolumn{1}{c|}{0.547} &
  {\color[HTML]{808080} 0.583} &
  \multicolumn{1}{c|}{0.580} &
  {\color[HTML]{808080} 0.939} &
  \multicolumn{1}{c|}{0.934} &
  {\color[HTML]{808080} 0.420} &
  \multicolumn{1}{c|}{0.415} &
  {\color[HTML]{808080} 0.695} &
  0.690 \\
\multicolumn{1}{c|}{} &
  DesTSeg &
  {\color[HTML]{808080} 0.964} &
  \multicolumn{1}{c|}{0.964} &
  {\color[HTML]{808080} 0.984} &
  \multicolumn{1}{c|}{0.986} &
  {\color[HTML]{808080} 0.961} &
  \multicolumn{1}{c|}{0.962} &
  {\color[HTML]{808080} 0.970} &
  0.971 &
  {\color[HTML]{808080} 0.932} &
  \multicolumn{1}{c|}{0.920} &
  {\color[HTML]{808080} 0.685} &
  \multicolumn{1}{c|}{\textbf{0.711}} &
  {\color[HTML]{808080} 0.657} &
  \multicolumn{1}{c|}{\textbf{0.682}} &
  {\color[HTML]{808080} 0.883} &
  \multicolumn{1}{c|}{0.834} &
  {\color[HTML]{808080} 0.501} &
  \multicolumn{1}{c|}{\textbf{0.528}} &
  {\color[HTML]{808080} 0.731} &
  0.735 \\
\multicolumn{1}{c|}{} &
  CFLOW-AD &
  {\color[HTML]{808080} 0.950} &
  \multicolumn{1}{c|}{0.916} &
  \multicolumn{1}{l}{{\color[HTML]{808080} 0.981}} &
  \multicolumn{1}{c|}{0.967} &
  \multicolumn{1}{l}{{\color[HTML]{808080} 0.953}} &
  \multicolumn{1}{c|}{0.934} &
  {\color[HTML]{808080} 0.961} &
  0.939 &
  \multicolumn{1}{l}{{\color[HTML]{808080} 0.967}} &
  \multicolumn{1}{c|}{0.957} &
  \multicolumn{1}{l}{{\color[HTML]{808080} 0.526}} &
  \multicolumn{1}{c|}{0.459} &
  \multicolumn{1}{l}{{\color[HTML]{808080} 0.556}} &
  \multicolumn{1}{c|}{0.486} &
  \multicolumn{1}{l}{{\color[HTML]{808080} 0.907}} &
  \multicolumn{1}{c|}{0.883} &
  \multicolumn{1}{l}{{\color[HTML]{808080} 0.396}} &
  \multicolumn{1}{c|}{0.332} &
  {\color[HTML]{808080} 0.670} &
  0.623 \\
\multicolumn{1}{c|}{} &
  PyramidFlow &
  {\color[HTML]{808080} 0.878} &
  \multicolumn{1}{c|}{0.702} &
  {\color[HTML]{808080} 0.944} &
  \multicolumn{1}{c|}{0.855} &
  {\color[HTML]{808080} 0.904} &
  \multicolumn{1}{c|}{0.855} &
  {\color[HTML]{808080} 0.909} &
  0.804 &
  {\color[HTML]{808080} 0.946} &
  \multicolumn{1}{c|}{0.800} &
  {\color[HTML]{808080} 0.463} &
  \multicolumn{1}{c|}{0.223} &
  {\color[HTML]{808080} 0.476} &
  \multicolumn{1}{c|}{0.220} &
  {\color[HTML]{808080} 0.820} &
  \multicolumn{1}{c|}{0.475} &
  {\color[HTML]{808080} 0.323} &
  \multicolumn{1}{c|}{0.128} &
  {\color[HTML]{808080} 0.605} &
  0.369 \\
\multicolumn{1}{c|}{} &
  CFA &
  {\color[HTML]{808080} 0.921} &
  \multicolumn{1}{c|}{0.576} &
  {\color[HTML]{808080} 0.966} &
  \multicolumn{1}{c|}{0.783} &
  {\color[HTML]{808080} 0.950} &
  \multicolumn{1}{c|}{0.847} &
  {\color[HTML]{808080} 0.946} &
  0.735 &
  {\color[HTML]{808080} 0.954} &
  \multicolumn{1}{c|}{0.548} &
  {\color[HTML]{808080} 0.476} &
  \multicolumn{1}{c|}{0.119} &
  {\color[HTML]{808080} 0.490} &
  \multicolumn{1}{c|}{0.147} &
  {\color[HTML]{808080} 0.856} &
  \multicolumn{1}{c|}{0.253} &
  {\color[HTML]{808080} 0.343} &
  \multicolumn{1}{c|}{0.089} &
  {\color[HTML]{808080} 0.624} &
  0.231 \\
\multicolumn{1}{c|}{\multirow{-10}{*}{One-class}} &
  PatchCore &
  {\color[HTML]{808080} 0.993} &
  \multicolumn{1}{c|}{\textbf{0.992}} &
  {\color[HTML]{808080} 0.998} &
  \multicolumn{1}{c|}{\textbf{0.998}} &
  {\color[HTML]{808080} 0.986} &
  \multicolumn{1}{c|}{{\ul 0.985}} &
  {\color[HTML]{808080} 0.992} &
  {\ul 0.992} &
  {\color[HTML]{808080} 0.981} &
  \multicolumn{1}{c|}{{\ul 0.980}} &
  {\color[HTML]{808080} 0.674} &
  \multicolumn{1}{c|}{0.661} &
  {\color[HTML]{808080} 0.660} &
  \multicolumn{1}{c|}{0.652} &
  {\color[HTML]{808080} 0.941} &
  \multicolumn{1}{c|}{0.940} &
  {\color[HTML]{808080} 0.500} &
  \multicolumn{1}{c|}{0.492} &
  {\color[HTML]{808080} 0.751} &
  0.745 \\ \midrule
\multicolumn{1}{c|}{} &
  UniAD &
  \multicolumn{1}{c}{-} &
  \multicolumn{1}{c|}{0.924} &
  - &
  \multicolumn{1}{c|}{0.973} &
  - &
  \multicolumn{1}{c|}{0.952} &
  - &
  0.950 &
  - &
  \multicolumn{1}{c|}{0.958} &
  - &
  \multicolumn{1}{c|}{0.429} &
  - &
  \multicolumn{1}{c|}{0.480} &
  - &
  \multicolumn{1}{c|}{0.887} &
  - &
  \multicolumn{1}{c|}{0.326} &
  - &
  0.616 \\
\multicolumn{1}{c|}{} &
  DiAD &
  \multicolumn{1}{c}{-} &
  \multicolumn{1}{c|}{0.889} &
  - &
  \multicolumn{1}{c|}{0.958} &
  - &
  \multicolumn{1}{c|}{0.935} &
  - &
  0.927 &
  - &
  \multicolumn{1}{c|}{0.893} &
  - &
  \multicolumn{1}{c|}{0.270} &
  - &
  \multicolumn{1}{c|}{0.325} &
  - &
  \multicolumn{1}{c|}{0.639} &
  - &
  \multicolumn{1}{c|}{0.211} &
  - &
  0.468 \\
\multicolumn{1}{c|}{} &
  ViTAD &
  \multicolumn{1}{c}{-} &
  \multicolumn{1}{c|}{{\ul 0.980}} &
  - &
  \multicolumn{1}{c|}{0.991} &
  - &
  \multicolumn{1}{c|}{0.968} &
  - &
  0.980 &
  - &
  \multicolumn{1}{c|}{0.977} &
  - &
  \multicolumn{1}{c|}{0.554} &
  - &
  \multicolumn{1}{c|}{0.586} &
  - &
  \multicolumn{1}{c|}{0.914} &
  - &
  \multicolumn{1}{c|}{0.425} &
  - &
  0.691 \\
\multicolumn{1}{c|}{} &
  InvAD &
  \multicolumn{1}{c}{-} &
  \multicolumn{1}{c|}{0.981} &
  - &
  \multicolumn{1}{c|}{0.991} &
  - &
  \multicolumn{1}{c|}{0.976} &
  - &
  0.983 &
  - &
  \multicolumn{1}{c|}{0.981} &
  - &
  \multicolumn{1}{c|}{0.562} &
  - &
  \multicolumn{1}{c|}{0.591} &
  - &
  \multicolumn{1}{c|}{\textbf{0.941}} &
  - &
  \multicolumn{1}{c|}{0.426} &
  - &
  0.700 \\
\multicolumn{1}{c|}{} &
  InvAD-lite &
  \multicolumn{1}{c}{-} &
  \multicolumn{1}{c|}{0.979} &
  - &
  \multicolumn{1}{c|}{{\ul 0.992}} &
  - &
  \multicolumn{1}{c|}{0.968} &
  - &
  0.980 &
  - &
  \multicolumn{1}{c|}{0.979} &
  - &
  \multicolumn{1}{c|}{0.544} &
  - &
  \multicolumn{1}{c|}{0.578} &
  - &
  \multicolumn{1}{c|}{0.933} &
  - &
  \multicolumn{1}{c|}{0.414} &
  - &
  0.690 \\
\multicolumn{1}{c|}{} &
  MambaAD &
  \multicolumn{1}{c}{-} &
  \multicolumn{1}{c|}{0.974} &
  - &
  \multicolumn{1}{c|}{0.991} &
  - &
  \multicolumn{1}{c|}{0.969} &
  - &
  0.978 &
  - &
  \multicolumn{1}{c|}{0.974} &
  - &
  \multicolumn{1}{c|}{0.545} &
  - &
  \multicolumn{1}{c|}{0.574} &
  - &
  \multicolumn{1}{c|}{0.932} &
  - &
  \multicolumn{1}{c|}{0.411} &
  - &
  0.687 \\
\multicolumn{1}{c|}{\multirow{-7}{*}{Multi-class}} &
  HierCore (Ours) &
  \multicolumn{1}{c}{-} &
  \multicolumn{1}{c|}{\textbf{0.992}} &
  - &
  \multicolumn{1}{c|}{\textbf{0.998}} &
  - &
  \multicolumn{1}{c|}{\textbf{0.986}} &
  - &
  \textbf{0.992} &
  - &
  \multicolumn{1}{c|}{\textbf{0.981}} &
  - &
  \multicolumn{1}{c|}{{\ul 0.668}} &
  - &
  \multicolumn{1}{c|}{{\ul 0.656}} &
  - &
  \multicolumn{1}{c|}{{\ul 0.940}} &
  - &
  \multicolumn{1}{c|}{{\ul 0.496}} &
  - &
  0.748 \\ \bottomrule
\end{tabular}
\end{adjustbox}
\end{table}

\begin{table}[H]
\caption{Anomaly detection performance on the VisA dataset. The table presents a comparison between one-class and multi-class unsupervised image anomaly detection models. One-class models are evaluated under both \textcolor{gray}{one-class} and multi-class settings. All experiments are conducted assuming unknown classes during training and known classes during evaluation.}\label{tab:exp_w_oc_mc_visa}
\vspace{5pt}
\centering
\begin{adjustbox}{max width=\textwidth}
\begin{tabular}{@{}cl|lccccccc|cccccccccc|cc@{}}
\toprule
\multicolumn{2}{c|}{} &
  \multicolumn{8}{c|}{Image-level} &
  \multicolumn{12}{c}{Pixel-level} \\ \cmidrule(l){3-22} 
\multicolumn{2}{c|}{\multirow{-2}{*}{Model}} &
  \multicolumn{2}{c|}{mAUROC} &
  \multicolumn{2}{c|}{mAP} &
  \multicolumn{2}{c|}{mF1-max} &
  \multicolumn{2}{c|}{mAD} &
  \multicolumn{2}{c|}{mAUROC} &
  \multicolumn{2}{c|}{mAP} &
  \multicolumn{2}{c|}{mF1-max} &
  \multicolumn{2}{c|}{mAUPRO} &
  \multicolumn{2}{c|}{mIoU-max} &
  \multicolumn{2}{c}{mAD} \\ \midrule
\multicolumn{1}{c|}{} &
  DRAEM &
  {\color[HTML]{808080} 0.727} &
  \multicolumn{1}{c|}{0.551} &
  {\color[HTML]{808080} 0.744} &
  \multicolumn{1}{c|}{0.624} &
  {\color[HTML]{808080} 0.759} &
  \multicolumn{1}{c|}{0.729} &
  {\color[HTML]{808080} 0.743} &
  0.635 &
  {\color[HTML]{808080} 0.488} &
  \multicolumn{1}{c|}{0.375} &
  {\color[HTML]{808080} 0.012} &
  \multicolumn{1}{c|}{0.006} &
  {\color[HTML]{808080} 0.036} &
  \multicolumn{1}{c|}{0.017} &
  {\color[HTML]{808080} 0.246} &
  \multicolumn{1}{c|}{0.100} &
  {\color[HTML]{808080} 0.019} &
  \multicolumn{1}{c|}{0.009} &
  {\color[HTML]{808080} 0.160} &
  0.101 \\
\multicolumn{1}{c|}{} &
  SimpleNet &
  {\color[HTML]{808080} 0.945} &
  \multicolumn{1}{c|}{0.864} &
  \multicolumn{1}{l}{{\color[HTML]{808080} 0.951}} &
  \multicolumn{1}{c|}{0.891} &
  \multicolumn{1}{l}{{\color[HTML]{808080} 0.910}} &
  \multicolumn{1}{c|}{0.828} &
  {\color[HTML]{808080} 0.936} &
  0.861 &
  \multicolumn{1}{l}{{\color[HTML]{808080} 0.980}} &
  \multicolumn{1}{c|}{0.966} &
  \multicolumn{1}{l}{{\color[HTML]{808080} 0.350}} &
  \multicolumn{1}{c|}{0.340} &
  \multicolumn{1}{l}{{\color[HTML]{808080} 0.395}} &
  \multicolumn{1}{c|}{0.378} &
  \multicolumn{1}{l}{{\color[HTML]{808080} 0.866}} &
  \multicolumn{1}{c|}{0.792} &
  \multicolumn{1}{l}{{\color[HTML]{808080} 0.268}} &
  \multicolumn{1}{c|}{0.257} &
  \multicolumn{1}{l}{{\color[HTML]{808080} 0.572}} &
  0.547 \\
\multicolumn{1}{c|}{} &
  RealNet &
  {\color[HTML]{808080} 0.922} &
  \multicolumn{1}{c|}{0.714} &
  {\color[HTML]{808080} 0.941} &
  \multicolumn{1}{c|}{0.795} &
  {\color[HTML]{808080} 0.888} &
  \multicolumn{1}{c|}{0.747} &
  {\color[HTML]{808080} 0.917} &
  0.752 &
  {\color[HTML]{808080} 0.864} &
  \multicolumn{1}{c|}{0.610} &
  {\color[HTML]{808080} 0.410} &
  \multicolumn{1}{c|}{0.257} &
  {\color[HTML]{808080} 0.454} &
  \multicolumn{1}{c|}{0.226} &
  {\color[HTML]{808080} 0.661} &
  \multicolumn{1}{c|}{0.274} &
  {\color[HTML]{808080} 0.313} &
  \multicolumn{1}{c|}{0.135} &
  \multicolumn{1}{l}{{\color[HTML]{808080} 0.540}} &
  0.300 \\
\multicolumn{1}{c|}{} &
  RD &
  {\color[HTML]{808080} 0.959} &
  \multicolumn{1}{c|}{0.906} &
  {\color[HTML]{808080} 0.962} &
  \multicolumn{1}{c|}{0.909} &
  {\color[HTML]{808080} 0.925} &
  \multicolumn{1}{c|}{0.893} &
  {\color[HTML]{808080} 0.949} &
  0.903 &
  {\color[HTML]{808080} 0.987} &
  \multicolumn{1}{c|}{0.980} &
  {\color[HTML]{808080} 0.409} &
  \multicolumn{1}{c|}{0.354} &
  {\color[HTML]{808080} 0.453} &
  \multicolumn{1}{c|}{0.425} &
  {\color[HTML]{808080} 0.934} &
  \multicolumn{1}{c|}{0.919} &
  {\color[HTML]{808080} 0.306} &
  \multicolumn{1}{c|}{0.279} &
  \multicolumn{1}{l}{{\color[HTML]{808080} 0.618}} &
  0.591 \\
\multicolumn{1}{c|}{} &
  RD++ &
  {\color[HTML]{808080} 0.952} &
  \multicolumn{1}{c|}{0.939} &
  {\color[HTML]{808080} 0.958} &
  \multicolumn{1}{c|}{0.947} &
  {\color[HTML]{808080} 0.917} &
  \multicolumn{1}{c|}{0.902} &
  {\color[HTML]{808080} 0.943} &
  0.929 &
  {\color[HTML]{808080} 0.988} &
  \multicolumn{1}{c|}{0.984} &
  {\color[HTML]{808080} 0.406} &
  \multicolumn{1}{c|}{0.423} &
  {\color[HTML]{808080} 0.450} &
  \multicolumn{1}{c|}{0.463} &
  {\color[HTML]{808080} 0.930} &
  \multicolumn{1}{c|}{0.919} &
  {\color[HTML]{808080} 0.305} &
  \multicolumn{1}{c|}{0.312} &
  \multicolumn{1}{l}{{\color[HTML]{808080} 0.616}} &
  0.620 \\
\multicolumn{1}{c|}{} &
  DesTSeg &
  {\color[HTML]{808080} 0.913} &
  \multicolumn{1}{c|}{0.899} &
  {\color[HTML]{808080} 0.923} &
  \multicolumn{1}{c|}{0.914} &
  {\color[HTML]{808080} 0.876} &
  \multicolumn{1}{c|}{0.867} &
  {\color[HTML]{808080} 0.904} &
  0.893 &
  {\color[HTML]{808080} 0.903} &
  \multicolumn{1}{c|}{0.867} &
  {\color[HTML]{808080} 0.367} &
  \multicolumn{1}{c|}{0.466} &
  {\color[HTML]{808080} 0.417} &
  \multicolumn{1}{c|}{0.472} &
  {\color[HTML]{808080} 0.703} &
  \multicolumn{1}{c|}{0.611} &
  {\color[HTML]{808080} 0.279} &
  \multicolumn{1}{c|}{0.327} &
  \multicolumn{1}{l}{{\color[HTML]{808080} 0.534}} &
  0.549 \\
\multicolumn{1}{c|}{} &
  CFLOW-AD &
  {\color[HTML]{808080} 0.901} &
  \multicolumn{1}{c|}{0.865} &
  \multicolumn{1}{l}{{\color[HTML]{808080} 0.910}} &
  \multicolumn{1}{c|}{0.888} &
  \multicolumn{1}{l}{{\color[HTML]{808080} 0.863}} &
  \multicolumn{1}{c|}{0.849} &
  {\color[HTML]{808080} 0.892} &
  0.867 &
  \multicolumn{1}{l}{{\color[HTML]{808080} 0.982}} &
  \multicolumn{1}{c|}{0.977} &
  \multicolumn{1}{l}{{\color[HTML]{808080} 0.329}} &
  \multicolumn{1}{c|}{0.339} &
  \multicolumn{1}{l}{{\color[HTML]{808080} 0.386}} &
  \multicolumn{1}{c|}{0.372} &
  \multicolumn{1}{l}{{\color[HTML]{808080} 0.884}} &
  \multicolumn{1}{c|}{0.868} &
  \multicolumn{1}{l}{{\color[HTML]{808080} 0.252}} &
  \multicolumn{1}{c|}{0.249} &
  \multicolumn{1}{l}{{\color[HTML]{808080} 0.567}} &
  0.561 \\
\multicolumn{1}{c|}{} &
  PyramidFlow &
  {\color[HTML]{808080} 0.902} &
  \multicolumn{1}{c|}{0.582} &
  {\color[HTML]{808080} 0.908} &
  \multicolumn{1}{c|}{0.663} &
  {\color[HTML]{808080} 0.867} &
  \multicolumn{1}{c|}{0.744} &
  {\color[HTML]{808080} 0.892} &
  0.663 &
  {\color[HTML]{808080} 0.952} &
  \multicolumn{1}{c|}{0.770} &
  {\color[HTML]{808080} 0.243} &
  \multicolumn{1}{c|}{0.072} &
  {\color[HTML]{808080} 0.312} &
  \multicolumn{1}{c|}{0.096} &
  {\color[HTML]{808080} 0.808} &
  \multicolumn{1}{c|}{0.428} &
  {\color[HTML]{808080} 0.191} &
  \multicolumn{1}{c|}{0.056} &
  \multicolumn{1}{l}{{\color[HTML]{808080} 0.501}} &
  0.284 \\
\multicolumn{1}{c|}{} &
  CFA &
  {\color[HTML]{808080} 0.861} &
  \multicolumn{1}{c|}{0.663} &
  {\color[HTML]{808080} 0.897} &
  \multicolumn{1}{c|}{0.743} &
  {\color[HTML]{808080} 0.847} &
  \multicolumn{1}{c|}{0.742} &
  {\color[HTML]{808080} 0.868} &
  0.716 &
  {\color[HTML]{808080} 0.956} &
  \multicolumn{1}{c|}{0.813} &
  {\color[HTML]{808080} 0.290} &
  \multicolumn{1}{c|}{0.221} &
  {\color[HTML]{808080} 0.341} &
  \multicolumn{1}{c|}{0.262} &
  {\color[HTML]{808080} 0.770} &
  \multicolumn{1}{c|}{0.508} &
  {\color[HTML]{808080} 0.223} &
  \multicolumn{1}{c|}{0.170} &
  \multicolumn{1}{l}{{\color[HTML]{808080} 0.516}} &
  0.395 \\
\multicolumn{1}{c|}{\multirow{-10}{*}{One-class}} &
  PatchCore &
  {\color[HTML]{808080} 0.963} &
  \multicolumn{1}{c|}{{\ul 0.961}} &
  {\color[HTML]{808080} 0.968} &
  \multicolumn{1}{c|}{{\ul 0.967}} &
  {\color[HTML]{808080} 0.927} &
  \multicolumn{1}{c|}{\textbf{0.930}} &
  {\color[HTML]{808080} 0.953} &
  {\ul 0.953} &
  {\color[HTML]{808080} 0.981} &
  \multicolumn{1}{c|}{0.980} &
  {\color[HTML]{808080} 0.505} &
  \multicolumn{1}{c|}{\textbf{0.514}} &
  {\color[HTML]{808080} 0.512} &
  \multicolumn{1}{c|}{\textbf{0.517}} &
  {\color[HTML]{808080} 0.909} &
  \multicolumn{1}{c|}{0.907} &
  {\color[HTML]{808080} 0.356} &
  \multicolumn{1}{c|}{{\ul 0.361}} &
  {\color[HTML]{808080} 0.653} &
  \textbf{0.656} \\ \midrule
\multicolumn{1}{c|}{} &
  UniAD &
  \multicolumn{1}{c}{-} &
  \multicolumn{1}{c|}{0.888} &
  - &
  \multicolumn{1}{c|}{0.908} &
  - &
  \multicolumn{1}{c|}{0.858} &
  - &
  0.884 &
  - &
  \multicolumn{1}{c|}{0.983} &
  - &
  \multicolumn{1}{c|}{0.337} &
  - &
  \multicolumn{1}{c|}{0.390} &
  - &
  \multicolumn{1}{c|}{0.855} &
  - &
  \multicolumn{1}{c|}{0.257} &
  - &
  0.564 \\
\multicolumn{1}{c|}{} &
  DiAD &
  \multicolumn{1}{c}{-} &
  \multicolumn{1}{c|}{0.848} &
  - &
  \multicolumn{1}{c|}{0.885} &
  - &
  \multicolumn{1}{c|}{0.869} &
  - &
  0.867 &
  - &
  \multicolumn{1}{c|}{0.825} &
  - &
  \multicolumn{1}{c|}{0.179} &
  - &
  \multicolumn{1}{c|}{0.232} &
  - &
  \multicolumn{1}{c|}{0.445} &
  - &
  \multicolumn{1}{c|}{0.149} &
  - &
  0.366 \\
\multicolumn{1}{c|}{} &
  ViTAD &
  \multicolumn{1}{c}{-} &
  \multicolumn{1}{c|}{0.905} &
  - &
  \multicolumn{1}{c|}{0.917} &
  - &
  \multicolumn{1}{c|}{0.863} &
  - &
  0.895 &
  - &
  \multicolumn{1}{c|}{0.982} &
  - &
  \multicolumn{1}{c|}{0.366} &
  - &
  \multicolumn{1}{c|}{0.411} &
  - &
  \multicolumn{1}{c|}{0.851} &
  - &
  \multicolumn{1}{c|}{0.276} &
  - &
  0.577 \\
\multicolumn{1}{c|}{} &
  InvAD &
  \multicolumn{1}{c}{-} &
  \multicolumn{1}{c|}{0.955} &
  - &
  \multicolumn{1}{c|}{0.958} &
  - &
  \multicolumn{1}{c|}{0.920} &
  - &
  0.944 &
  - &
  \multicolumn{1}{c|}{0.989} &
  - &
  \multicolumn{1}{c|}{0.431} &
  - &
  \multicolumn{1}{c|}{0.470} &
  - &
  \multicolumn{1}{c|}{{\ul 0.926}} &
  - &
  \multicolumn{1}{c|}{0.327} &
  - &
  0.628 \\
\multicolumn{1}{c|}{} &
  InvAD-lite &
  \multicolumn{1}{c}{-} &
  \multicolumn{1}{c|}{0.949} &
  - &
  \multicolumn{1}{c|}{0.952} &
  - &
  \multicolumn{1}{c|}{0.907} &
  - &
  0.936 &
  - &
  \multicolumn{1}{c|}{{\ul 0.986}} &
  - &
  \multicolumn{1}{c|}{0.402} &
  - &
  \multicolumn{1}{c|}{0.440} &
  - &
  \multicolumn{1}{c|}{\textbf{0.931}} &
  - &
  \multicolumn{1}{c|}{0.298} &
  - &
  0.611 \\
\multicolumn{1}{c|}{} &
  MambaAD &
  \multicolumn{1}{c}{-} &
  \multicolumn{1}{c|}{0.941} &
  - &
  \multicolumn{1}{c|}{0.948} &
  - &
  \multicolumn{1}{c|}{0.893} &
  - &
  0.927 &
  - &
  \multicolumn{1}{c|}{0.985} &
  - &
  \multicolumn{1}{c|}{0.395} &
  - &
  \multicolumn{1}{c|}{0.437} &
  - &
  \multicolumn{1}{c|}{0.914} &
  - &
  \multicolumn{1}{c|}{0.295} &
  - &
  0.605 \\
\multicolumn{1}{c|}{\multirow{-7}{*}{Multi-class}} &
  HierCore (Ours) &
  \multicolumn{1}{c}{-} &
  \multicolumn{1}{c|}{\textbf{0.963}} &
  - &
  \multicolumn{1}{c|}{\textbf{0.968}} &
  - &
  \multicolumn{1}{c|}{{\ul 0.927}} &
  - &
  \textbf{0.953} &
  - &
  \multicolumn{1}{c|}{0.981} &
  - &
  \multicolumn{1}{c|}{{\ul 0.498}} &
  - &
  \multicolumn{1}{c|}{{\ul 0.507}} &
  - &
  \multicolumn{1}{c|}{0.909} &
  - &
  \multicolumn{1}{c|}{\textbf{0.350}} &
  - &
  {\ul 0.649} \\ \bottomrule
\end{tabular}
\end{adjustbox}
\end{table}

\begin{table}[H]
\caption{Anomaly detection performance on the MPDD dataset. The table presents a comparison between one-class and multi-class unsupervised image anomaly detection models. One-class models are evaluated under both \textcolor{gray}{one-class} and multi-class settings. All experiments are conducted assuming unknown classes during training and known classes during evaluation.}\label{tab:exp_w_oc_mc_mpdd}
\vspace{5pt}
\centering
\begin{adjustbox}{max width=\textwidth}
\begin{tabular}{@{}cl|lccccccc|cccccccccc|cc@{}}
\toprule
\multicolumn{2}{c|}{} &
  \multicolumn{8}{c|}{Image-level} &
  \multicolumn{12}{c}{Pixel-level} \\ \cmidrule(l){3-22} 
\multicolumn{2}{c|}{\multirow{-2}{*}{Model}} &
  \multicolumn{2}{c|}{mAUROC} &
  \multicolumn{2}{c|}{mAP} &
  \multicolumn{2}{c|}{mF1-max} &
  \multicolumn{2}{c|}{mAD} &
  \multicolumn{2}{c|}{mAUROC} &
  \multicolumn{2}{c|}{mAP} &
  \multicolumn{2}{c|}{mF1-max} &
  \multicolumn{2}{c|}{mAUPRO} &
  \multicolumn{2}{c|}{mIoU-max} &
  \multicolumn{2}{c}{mAD} \\ \midrule
\multicolumn{1}{c|}{} &
  DRAEM &
  {\color[HTML]{808080} 0.706} &
  \multicolumn{1}{c|}{0.356} &
  {\color[HTML]{808080} 0.785} &
  \multicolumn{1}{c|}{0.533} &
  {\color[HTML]{808080} 0.842} &
  \multicolumn{1}{c|}{0.725} &
  {\color[HTML]{808080} 0.778} &
  0.538 &
  {\color[HTML]{808080} 0.427} &
  \multicolumn{1}{c|}{0.421} &
  {\color[HTML]{808080} 0.019} &
  \multicolumn{1}{c|}{0.022} &
  {\color[HTML]{808080} 0.053} &
  \multicolumn{1}{c|}{0.046} &
  {\color[HTML]{808080} 0.242} &
  \multicolumn{1}{c|}{0.190} &
  {\color[HTML]{808080} 0.029} &
  \multicolumn{1}{c|}{0.025} &
  {\color[HTML]{808080} 0.154} &
  0.141 \\
\multicolumn{1}{c|}{} &
  SimpleNet &
  {\color[HTML]{808080} 0.954} &
  \multicolumn{1}{c|}{0.884} &
  \multicolumn{1}{l}{{\color[HTML]{808080} 0.957}} &
  \multicolumn{1}{c|}{0.920} &
  \multicolumn{1}{l}{{\color[HTML]{808080} 0.940}} &
  \multicolumn{1}{c|}{0.879} &
  {\color[HTML]{808080} 0.950} &
  0.894 &
  \multicolumn{1}{l}{{\color[HTML]{808080} 0.938}} &
  \multicolumn{1}{c|}{0.965} &
  \multicolumn{1}{l}{{\color[HTML]{808080} 0.314}} &
  \multicolumn{1}{c|}{0.320} &
  \multicolumn{1}{l}{{\color[HTML]{808080} 0.340}} &
  \multicolumn{1}{c|}{0.346} &
  \multicolumn{1}{l}{{\color[HTML]{808080} 0.788}} &
  \multicolumn{1}{c|}{0.890} &
  \multicolumn{1}{l}{{\color[HTML]{808080} 0.240}} &
  \multicolumn{1}{c|}{0.245} &
  \multicolumn{1}{l}{{\color[HTML]{808080} 0.524}} &
  0.553 \\
\multicolumn{1}{c|}{} &
  RealNet &
  {\color[HTML]{808080} 0.764} &
  \multicolumn{1}{c|}{0.851} &
  {\color[HTML]{808080} 0.810} &
  \multicolumn{1}{c|}{0.902} &
  {\color[HTML]{808080} 0.803} &
  \multicolumn{1}{c|}{0.883} &
  {\color[HTML]{808080} 0.792} &
  0.879 &
  {\color[HTML]{808080} 0.876} &
  \multicolumn{1}{c|}{0.833} &
  {\color[HTML]{808080} 0.229} &
  \multicolumn{1}{c|}{0.361} &
  {\color[HTML]{808080} 0.267} &
  \multicolumn{1}{c|}{0.396} &
  {\color[HTML]{808080} 0.731} &
  \multicolumn{1}{c|}{0.681} &
  {\color[HTML]{808080} 0.186} &
  \multicolumn{1}{c|}{0.282} &
  \multicolumn{1}{l}{{\color[HTML]{808080} 0.458}} &
  0.511 \\
\multicolumn{1}{c|}{} &
  RD &
  {\color[HTML]{808080} 0.944} &
  \multicolumn{1}{c|}{0.913} &
  {\color[HTML]{808080} 0.965} &
  \multicolumn{1}{c|}{0.936} &
  {\color[HTML]{808080} 0.922} &
  \multicolumn{1}{c|}{0.918} &
  {\color[HTML]{808080} 0.944} &
  0.922 &
  {\color[HTML]{808080} 0.977} &
  \multicolumn{1}{c|}{0.983} &
  {\color[HTML]{808080} 0.432} &
  \multicolumn{1}{c|}{0.404} &
  {\color[HTML]{808080} 0.461} &
  \multicolumn{1}{c|}{0.418} &
  {\color[HTML]{808080} 0.936} &
  \multicolumn{1}{c|}{\textbf{0.955}} &
  {\color[HTML]{808080} 0.343} &
  \multicolumn{1}{c|}{0.314} &
  \multicolumn{1}{l}{{\color[HTML]{808080} 0.630}} &
  0.615 \\
\multicolumn{1}{c|}{} &
  RD++ &
  {\color[HTML]{808080} 0.918} &
  \multicolumn{1}{c|}{0.902} &
  {\color[HTML]{808080} 0.932} &
  \multicolumn{1}{c|}{0.933} &
  {\color[HTML]{808080} 0.914} &
  \multicolumn{1}{c|}{0.905} &
  {\color[HTML]{808080} 0.921} &
  0.913 &
  {\color[HTML]{808080} 0.979} &
  \multicolumn{1}{c|}{0.985} &
  {\color[HTML]{808080} 0.409} &
  \multicolumn{1}{c|}{0.430} &
  {\color[HTML]{808080} 0.429} &
  \multicolumn{1}{c|}{0.441} &
  {\color[HTML]{808080} 0.940} &
  \multicolumn{1}{c|}{\textbf{0.955}} &
  {\color[HTML]{808080} 0.314} &
  \multicolumn{1}{c|}{0.336} &
  \multicolumn{1}{l}{{\color[HTML]{808080} 0.614}} &
  0.629 \\
\multicolumn{1}{c|}{} &
  DesTSeg &
  {\color[HTML]{808080} 0.918} &
  \multicolumn{1}{c|}{0.913} &
  {\color[HTML]{808080} 0.938} &
  \multicolumn{1}{c|}{0.908} &
  {\color[HTML]{808080} 0.907} &
  \multicolumn{1}{c|}{0.902} &
  {\color[HTML]{808080} 0.921} &
  0.908 &
  {\color[HTML]{808080} 0.886} &
  \multicolumn{1}{c|}{0.820} &
  {\color[HTML]{808080} 0.363} &
  \multicolumn{1}{c|}{0.326} &
  {\color[HTML]{808080} 0.380} &
  \multicolumn{1}{c|}{0.346} &
  {\color[HTML]{808080} 0.734} &
  \multicolumn{1}{c|}{0.633} &
  {\color[HTML]{808080} 0.284} &
  \multicolumn{1}{c|}{0.256} &
  \multicolumn{1}{l}{{\color[HTML]{808080} 0.529}} &
  0.476 \\
\multicolumn{1}{c|}{} &
  CFLOW-AD &
  {\color[HTML]{808080} 0.850} &
  \multicolumn{1}{c|}{0.757} &
  \multicolumn{1}{l}{{\color[HTML]{808080} 0.880}} &
  \multicolumn{1}{c|}{0.801} &
  \multicolumn{1}{l}{{\color[HTML]{808080} 0.865}} &
  \multicolumn{1}{c|}{0.817} &
  {\color[HTML]{808080} 0.865} &
  0.792 &
  \multicolumn{1}{l}{{\color[HTML]{808080} 0.977}} &
  \multicolumn{1}{c|}{0.968} &
  \multicolumn{1}{l}{{\color[HTML]{808080} 0.339}} &
  \multicolumn{1}{c|}{0.263} &
  \multicolumn{1}{l}{{\color[HTML]{808080} 0.341}} &
  \multicolumn{1}{c|}{0.280} &
  \multicolumn{1}{l}{{\color[HTML]{808080} 0.913}} &
  \multicolumn{1}{c|}{0.895} &
  \multicolumn{1}{l}{{\color[HTML]{808080} 0.248}} &
  \multicolumn{1}{c|}{0.201} &
  \multicolumn{1}{l}{{\color[HTML]{808080} 0.564}} &
  0.521 \\
\multicolumn{1}{c|}{} &
  PyramidFlow &
  {\color[HTML]{808080} 0.843} &
  \multicolumn{1}{c|}{0.736} &
  {\color[HTML]{808080} 0.865} &
  \multicolumn{1}{c|}{0.770} &
  {\color[HTML]{808080} 0.843} &
  \multicolumn{1}{c|}{0.794} &
  {\color[HTML]{808080} 0.850} &
  0.767 &
  {\color[HTML]{808080} 0.973} &
  \multicolumn{1}{c|}{0.941} &
  {\color[HTML]{808080} 0.291} &
  \multicolumn{1}{c|}{0.211} &
  {\color[HTML]{808080} 0.311} &
  \multicolumn{1}{c|}{0.178} &
  {\color[HTML]{808080} 0.883} &
  \multicolumn{1}{c|}{0.772} &
  {\color[HTML]{808080} 0.218} &
  \multicolumn{1}{c|}{0.104} &
  \multicolumn{1}{l}{{\color[HTML]{808080} 0.535}} &
  0.441 \\
\multicolumn{1}{c|}{} &
  CFA &
  {\color[HTML]{808080} 0.866} &
  \multicolumn{1}{c|}{0.816} &
  {\color[HTML]{808080} 0.872} &
  \multicolumn{1}{c|}{0.877} &
  {\color[HTML]{808080} 0.879} &
  \multicolumn{1}{c|}{0.857} &
  {\color[HTML]{808080} 0.872} &
  0.850 &
  {\color[HTML]{808080} 0.930} &
  \multicolumn{1}{c|}{0.849} &
  {\color[HTML]{808080} 0.196} &
  \multicolumn{1}{c|}{0.196} &
  {\color[HTML]{808080} 0.217} &
  \multicolumn{1}{c|}{0.229} &
  {\color[HTML]{808080} 0.701} &
  \multicolumn{1}{c|}{0.535} &
  {\color[HTML]{808080} 0.161} &
  \multicolumn{1}{c|}{0.166} &
  \multicolumn{1}{l}{{\color[HTML]{808080} 0.441}} &
  0.395 \\
\multicolumn{1}{c|}{\multirow{-10}{*}{One-class}} &
  PatchCore &
  {\color[HTML]{808080} 0.945} &
  \multicolumn{1}{c|}{\textbf{0.948}} &
  {\color[HTML]{808080} 0.968} &
  \multicolumn{1}{c|}{\textbf{0.969}} &
  {\color[HTML]{808080} 0.933} &
  \multicolumn{1}{c|}{\textbf{0.938}} &
  {\color[HTML]{808080} 0.949} &
  {\ul 0.952} &
  {\color[HTML]{808080} 0.985} &
  \multicolumn{1}{c|}{\textbf{0.987}} &
  {\color[HTML]{808080} 0.479} &
  \multicolumn{1}{c|}{\textbf{0.484}} &
  {\color[HTML]{808080} 0.501} &
  \multicolumn{1}{c|}{\textbf{0.504}} &
  {\color[HTML]{808080} 0.945} &
  \multicolumn{1}{c|}{{\ul 0.950}} &
  {\color[HTML]{808080} 0.368} &
  \multicolumn{1}{c|}{\textbf{0.371}} &
  {\color[HTML]{808080} 0.656} &
  \textbf{0.659} \\ \midrule
\multicolumn{1}{c|}{} &
  UniAD &
  \multicolumn{1}{c}{-} &
  \multicolumn{1}{c|}{0.707} &
  - &
  \multicolumn{1}{c|}{0.754} &
  - &
  \multicolumn{1}{c|}{0.794} &
  - &
  0.752 &
  - &
  \multicolumn{1}{c|}{0.942} &
  - &
  \multicolumn{1}{c|}{0.139} &
  - &
  \multicolumn{1}{c|}{0.203} &
  - &
  \multicolumn{1}{c|}{0.799} &
  - &
  \multicolumn{1}{c|}{0.128} &
  - &
  0.442 \\
\multicolumn{1}{c|}{} &
  DiAD &
  \multicolumn{1}{c}{-} &
  \multicolumn{1}{c|}{0.683} &
  - &
  \multicolumn{1}{c|}{0.779} &
  - &
  \multicolumn{1}{c|}{0.801} &
  - &
  0.754 &
  - &
  \multicolumn{1}{c|}{0.904} &
  - &
  \multicolumn{1}{c|}{0.109} &
  - &
  \multicolumn{1}{c|}{0.131} &
  - &
  \multicolumn{1}{c|}{0.661} &
  - &
  \multicolumn{1}{c|}{0.082} &
  - &
  0.377 \\
\multicolumn{1}{c|}{} &
  ViTAD &
  \multicolumn{1}{c}{-} &
  \multicolumn{1}{c|}{0.870} &
  - &
  \multicolumn{1}{c|}{0.901} &
  - &
  \multicolumn{1}{c|}{0.872} &
  - &
  0.881 &
  - &
  \multicolumn{1}{c|}{0.977} &
  - &
  \multicolumn{1}{c|}{0.354} &
  - &
  \multicolumn{1}{c|}{0.375} &
  - &
  \multicolumn{1}{c|}{0.926} &
  - &
  \multicolumn{1}{c|}{0.278} &
  - &
  0.582 \\
\multicolumn{1}{c|}{} &
  InvAD &
  \multicolumn{1}{c}{-} &
  \multicolumn{1}{c|}{0.931} &
  - &
  \multicolumn{1}{c|}{0.943} &
  - &
  \multicolumn{1}{c|}{0.928} &
  - &
  0.934 &
  - &
  \multicolumn{1}{c|}{0.983} &
  - &
  \multicolumn{1}{c|}{0.428} &
  - &
  \multicolumn{1}{c|}{0.460} &
  - &
  \multicolumn{1}{c|}{0.947} &
  - &
  \multicolumn{1}{c|}{0.341} &
  - &
  0.632 \\
\multicolumn{1}{c|}{} &
  InvAD-lite &
  \multicolumn{1}{c}{-} &
  \multicolumn{1}{c|}{0.909} &
  - &
  \multicolumn{1}{c|}{0.929} &
  - &
  \multicolumn{1}{c|}{0.895} &
  - &
  0.911 &
  - &
  \multicolumn{1}{c|}{0.980} &
  - &
  \multicolumn{1}{c|}{0.397} &
  - &
  \multicolumn{1}{c|}{0.426} &
  - &
  \multicolumn{1}{c|}{0.940} &
  - &
  \multicolumn{1}{c|}{0.309} &
  - &
  0.610 \\
\multicolumn{1}{c|}{} &
  MambaAD &
  \multicolumn{1}{c}{-} &
  \multicolumn{1}{c|}{0.877} &
  - &
  \multicolumn{1}{c|}{0.927} &
  - &
  \multicolumn{1}{c|}{0.905} &
  - &
  0.903 &
  - &
  \multicolumn{1}{c|}{0.976} &
  - &
  \multicolumn{1}{c|}{0.349} &
  - &
  \multicolumn{1}{c|}{0.394} &
  - &
  \multicolumn{1}{c|}{0.926} &
  - &
  \multicolumn{1}{c|}{0.278} &
  - &
  0.585 \\
\multicolumn{1}{c|}{\multirow{-7}{*}{Multi-class}} &
  HierCore (Ours) &
  \multicolumn{1}{c}{-} &
  \multicolumn{1}{c|}{{\ul 0.943}} &
  - &
  \multicolumn{1}{c|}{{\ul 0.967}} &
  - &
  \multicolumn{1}{c|}{{\ul 0.934}} &
  - &
  \textbf{0.948} &
  - &
  \multicolumn{1}{c|}{{\ul 0.985}} &
  - &
  \multicolumn{1}{c|}{{\ul 0.477}} &
  - &
  \multicolumn{1}{c|}{{\ul 0.500}} &
  - &
  \multicolumn{1}{c|}{0.945} &
  - &
  \multicolumn{1}{c|}{{\ul 0.368}} &
  - &
  {\ul 0.655} \\ \bottomrule
\end{tabular}
\end{adjustbox}
\end{table}

\begin{table}[H]
\caption{Anomaly detection performance on the BTAD dataset. The table presents a comparison between one-class and multi-class unsupervised image anomaly detection models. One-class models are evaluated under both \textcolor{gray}{one-class} and multi-class settings. All experiments are conducted assuming unknown classes during training and known classes during evaluation.}\label{tab:exp_w_oc_mc_btad}
\vspace{5pt}
\centering
\begin{adjustbox}{max width=\textwidth}
\begin{tabular}{@{}cl|lccccccc|cccccccccc|cc@{}}
\toprule
\multicolumn{2}{c|}{} &
  \multicolumn{8}{c|}{Image-level} &
  \multicolumn{12}{c}{Pixel-level} \\ \cmidrule(l){3-22} 
\multicolumn{2}{c|}{\multirow{-2}{*}{Model}} &
  \multicolumn{2}{c|}{mAUROC} &
  \multicolumn{2}{c|}{mAP} &
  \multicolumn{2}{c|}{mF1-max} &
  \multicolumn{2}{c|}{mAD} &
  \multicolumn{2}{c|}{mAUROC} &
  \multicolumn{2}{c|}{mAP} &
  \multicolumn{2}{c|}{mF1-max} &
  \multicolumn{2}{c|}{mAUPRO} &
  \multicolumn{2}{c|}{mIoU-max} &
  \multicolumn{2}{c}{mAD} \\ \midrule
\multicolumn{1}{c|}{} &
  DRAEM &
  {\color[HTML]{808080} 0.717} &
  \multicolumn{1}{c|}{0.713} &
  {\color[HTML]{808080} 0.762} &
  \multicolumn{1}{c|}{0.785} &
  {\color[HTML]{808080} 0.781} &
  \multicolumn{1}{c|}{0.780} &
  {\color[HTML]{808080} 0.754} &
  0.759 &
  {\color[HTML]{808080} 0.502} &
  \multicolumn{1}{c|}{0.490} &
  {\color[HTML]{808080} 0.110} &
  \multicolumn{1}{c|}{0.037} &
  {\color[HTML]{808080} 0.086} &
  \multicolumn{1}{c|}{0.065} &
  {\color[HTML]{808080} 0.169} &
  \multicolumn{1}{c|}{0.162} &
  {\color[HTML]{808080} 0.045} &
  \multicolumn{1}{c|}{0.034} &
  {\color[HTML]{808080} 0.182} &
  0.158 \\
\multicolumn{1}{c|}{} &
  SimpleNet &
  {\color[HTML]{808080} 0.936} &
  \multicolumn{1}{c|}{0.932} &
  \multicolumn{1}{l}{{\color[HTML]{808080} 0.978}} &
  \multicolumn{1}{c|}{0.973} &
  \multicolumn{1}{l}{{\color[HTML]{808080} 0.942}} &
  \multicolumn{1}{c|}{0.933} &
  {\color[HTML]{808080} 0.952} &
  0.946 &
  \multicolumn{1}{l}{{\color[HTML]{808080} 0.965}} &
  \multicolumn{1}{c|}{0.963} &
  \multicolumn{1}{l}{{\color[HTML]{808080} 0.458}} &
  \multicolumn{1}{c|}{0.415} &
  \multicolumn{1}{l}{{\color[HTML]{808080} 0.471}} &
  \multicolumn{1}{c|}{0.443} &
  \multicolumn{1}{l}{{\color[HTML]{808080} 0.704}} &
  \multicolumn{1}{c|}{0.698} &
  \multicolumn{1}{l}{{\color[HTML]{808080} 0.310}} &
  \multicolumn{1}{c|}{0.286} &
  \multicolumn{1}{l}{{\color[HTML]{808080} 0.582}} &
  0.561 \\
\multicolumn{1}{c|}{} &
  RealNet &
  {\color[HTML]{808080} 0.950} &
  \multicolumn{1}{c|}{0.897} &
  {\color[HTML]{808080} 0.969} &
  \multicolumn{1}{c|}{0.953} &
  {\color[HTML]{808080} 0.928} &
  \multicolumn{1}{c|}{0.928} &
  {\color[HTML]{808080} 0.949} &
  0.926 &
  {\color[HTML]{808080} 0.954} &
  \multicolumn{1}{c|}{0.840} &
  {\color[HTML]{808080} 0.538} &
  \multicolumn{1}{c|}{0.481} &
  {\color[HTML]{808080} 0.570} &
  \multicolumn{1}{c|}{0.527} &
  {\color[HTML]{808080} 0.750} &
  \multicolumn{1}{c|}{0.534} &
  {\color[HTML]{808080} 0.401} &
  \multicolumn{1}{c|}{0.366} &
  \multicolumn{1}{l}{{\color[HTML]{808080} 0.643}} &
  0.550 \\
\multicolumn{1}{c|}{} &
  RD &
  {\color[HTML]{808080} 0.943} &
  \multicolumn{1}{c|}{0.944} &
  {\color[HTML]{808080} 0.952} &
  \multicolumn{1}{c|}{0.966} &
  {\color[HTML]{808080} 0.929} &
  \multicolumn{1}{c|}{0.940} &
  {\color[HTML]{808080} 0.941} &
  0.950 &
  {\color[HTML]{808080} 0.980} &
  \multicolumn{1}{c|}{\textbf{0.981}} &
  {\color[HTML]{808080} 0.583} &
  \multicolumn{1}{c|}{0.596} &
  {\color[HTML]{808080} 0.583} &
  \multicolumn{1}{c|}{0.592} &
  {\color[HTML]{808080} 0.801} &
  \multicolumn{1}{c|}{\textbf{0.807}} &
  {\color[HTML]{808080} 0.413} &
  \multicolumn{1}{c|}{0.421} &
  \multicolumn{1}{l}{{\color[HTML]{808080} 0.672}} &
  {\ul 0.679} \\
\multicolumn{1}{c|}{} &
  RD++ &
  {\color[HTML]{808080} 0.947} &
  \multicolumn{1}{c|}{0.946} &
  {\color[HTML]{808080} 0.969} &
  \multicolumn{1}{c|}{{\ul 0.978}} &
  {\color[HTML]{808080} 0.936} &
  \multicolumn{1}{c|}{0.941} &
  {\color[HTML]{808080} 0.951} &
  0.955 &
  {\color[HTML]{808080} 0.980} &
  \multicolumn{1}{c|}{{\ul 0.980}} &
  {\color[HTML]{808080} 0.580} &
  \multicolumn{1}{c|}{0.596} &
  {\color[HTML]{808080} 0.588} &
  \multicolumn{1}{c|}{0.598} &
  {\color[HTML]{808080} 0.795} &
  \multicolumn{1}{c|}{0.790} &
  {\color[HTML]{808080} 0.418} &
  \multicolumn{1}{c|}{{\ul 0.428}} &
  \multicolumn{1}{l}{{\color[HTML]{808080} 0.672}} &
  0.678 \\
\multicolumn{1}{c|}{} &
  DesTSeg &
  {\color[HTML]{808080} 0.885} &
  \multicolumn{1}{c|}{0.928} &
  {\color[HTML]{808080} 0.941} &
  \multicolumn{1}{c|}{0.959} &
  {\color[HTML]{808080} 0.918} &
  \multicolumn{1}{c|}{0.923} &
  {\color[HTML]{808080} 0.915} &
  0.937 &
  {\color[HTML]{808080} 0.944} &
  \multicolumn{1}{c|}{0.922} &
  {\color[HTML]{808080} 0.376} &
  \multicolumn{1}{c|}{0.348} &
  {\color[HTML]{808080} 0.419} &
  \multicolumn{1}{c|}{0.443} &
  {\color[HTML]{808080} 0.728} &
  \multicolumn{1}{c|}{0.700} &
  {\color[HTML]{808080} 0.268} &
  \multicolumn{1}{c|}{0.290} &
  \multicolumn{1}{l}{{\color[HTML]{808080} 0.547}} &
  0.541 \\
\multicolumn{1}{c|}{} &
  CFLOW-AD &
  {\color[HTML]{808080} 0.934} &
  \multicolumn{1}{c|}{0.912} &
  \multicolumn{1}{l}{{\color[HTML]{808080} 0.972}} &
  \multicolumn{1}{c|}{0.948} &
  \multicolumn{1}{l}{{\color[HTML]{808080} 0.935}} &
  \multicolumn{1}{c|}{0.883} &
  {\color[HTML]{808080} 0.947} &
  0.914 &
  \multicolumn{1}{l}{{\color[HTML]{808080} 0.970}} &
  \multicolumn{1}{c|}{0.968} &
  \multicolumn{1}{l}{{\color[HTML]{808080} 0.492}} &
  \multicolumn{1}{c|}{0.456} &
  \multicolumn{1}{l}{{\color[HTML]{808080} 0.418}} &
  \multicolumn{1}{c|}{0.501} &
  \multicolumn{1}{l}{{\color[HTML]{808080} 0.749}} &
  \multicolumn{1}{c|}{0.727} &
  \multicolumn{1}{l}{{\color[HTML]{808080} 0.278}} &
  \multicolumn{1}{c|}{0.338} &
  \multicolumn{1}{l}{{\color[HTML]{808080} 0.581}} &
  0.598 \\
\multicolumn{1}{c|}{} &
  PyramidFlow &
  {\color[HTML]{808080} 0.864} &
  \multicolumn{1}{c|}{0.870} &
  {\color[HTML]{808080} 0.756} &
  \multicolumn{1}{c|}{0.831} &
  {\color[HTML]{808080} 0.772} &
  \multicolumn{1}{c|}{0.810} &
  {\color[HTML]{808080} 0.797} &
  0.837 &
  {\color[HTML]{808080} 0.927} &
  \multicolumn{1}{c|}{0.909} &
  {\color[HTML]{808080} 0.395} &
  \multicolumn{1}{c|}{0.296} &
  {\color[HTML]{808080} 0.351} &
  \multicolumn{1}{c|}{0.269} &
  {\color[HTML]{808080} 0.694} &
  \multicolumn{1}{c|}{0.641} &
  {\color[HTML]{808080} 0.249} &
  \multicolumn{1}{c|}{0.183} &
  \multicolumn{1}{l}{{\color[HTML]{808080} 0.523}} &
  0.460 \\
\multicolumn{1}{c|}{} &
  CFA &
  {\color[HTML]{808080} 0.931} &
  \multicolumn{1}{c|}{0.927} &
  {\color[HTML]{808080} 0.981} &
  \multicolumn{1}{c|}{0.975} &
  {\color[HTML]{808080} 0.953} &
  \multicolumn{1}{c|}{0.935} &
  {\color[HTML]{808080} 0.955} &
  0.946 &
  {\color[HTML]{808080} 0.972} &
  \multicolumn{1}{c|}{0.963} &
  {\color[HTML]{808080} 0.569} &
  \multicolumn{1}{c|}{0.474} &
  {\color[HTML]{808080} 0.562} &
  \multicolumn{1}{c|}{0.502} &
  {\color[HTML]{808080} 0.746} &
  \multicolumn{1}{c|}{0.695} &
  {\color[HTML]{808080} 0.391} &
  \multicolumn{1}{c|}{0.336} &
  \multicolumn{1}{l}{{\color[HTML]{808080} 0.648}} &
  0.594 \\
\multicolumn{1}{c|}{\multirow{-10}{*}{One-class}} &
  PatchCore &
  {\color[HTML]{808080} 0.941} &
  \multicolumn{1}{c|}{0.944} &
  {\color[HTML]{808080} 0.978} &
  \multicolumn{1}{c|}{{\ul 0.978}} &
  {\color[HTML]{808080} 0.938} &
  \multicolumn{1}{c|}{0.947} &
  {\color[HTML]{808080} 0.952} &
  0.956 &
  {\color[HTML]{808080} 0.972} &
  \multicolumn{1}{c|}{0.972} &
  {\color[HTML]{808080} 0.580} &
  \multicolumn{1}{c|}{0.582} &
  {\color[HTML]{808080} 0.563} &
  \multicolumn{1}{c|}{0.564} &
  {\color[HTML]{808080} 0.753} &
  \multicolumn{1}{c|}{0.753} &
  {\color[HTML]{808080} 0.393} &
  \multicolumn{1}{c|}{0.394} &
  {\color[HTML]{808080} 0.652} &
  0.653 \\ \midrule
\multicolumn{1}{c|}{} &
  UniAD &
  \multicolumn{1}{c}{-} &
  \multicolumn{1}{c|}{{\ul 0.949}} &
  - &
  \multicolumn{1}{c|}{\textbf{0.983}} &
  - &
  \multicolumn{1}{c|}{{\ul 0.948}} &
  - &
  {\ul 0.960} &
  - &
  \multicolumn{1}{c|}{0.972} &
  - &
  \multicolumn{1}{c|}{0.501} &
  - &
  \multicolumn{1}{c|}{0.537} &
  - &
  \multicolumn{1}{c|}{0.779} &
  - &
  \multicolumn{1}{c|}{0.368} &
  - &
  0.631 \\
\multicolumn{1}{c|}{} &
  DiAD &
  \multicolumn{1}{c}{-} &
  \multicolumn{1}{c|}{0.901} &
  - &
  \multicolumn{1}{c|}{0.884} &
  - &
  \multicolumn{1}{c|}{0.926} &
  - &
  0.904 &
  - &
  \multicolumn{1}{c|}{0.917} &
  - &
  \multicolumn{1}{c|}{0.196} &
  - &
  \multicolumn{1}{c|}{0.267} &
  - &
  \multicolumn{1}{c|}{0.704} &
  - &
  \multicolumn{1}{c|}{0.157} &
  - &
  0.448 \\
\multicolumn{1}{c|}{} &
  ViTAD &
  \multicolumn{1}{c}{-} &
  \multicolumn{1}{c|}{0.940} &
  - &
  \multicolumn{1}{c|}{0.972} &
  - &
  \multicolumn{1}{c|}{0.938} &
  - &
  0.950 &
  - &
  \multicolumn{1}{c|}{0.975} &
  - &
  \multicolumn{1}{c|}{\textbf{0.639}} &
  - &
  \multicolumn{1}{c|}{{\ul 0.600}} &
  - &
  \multicolumn{1}{c|}{0.711} &
  - &
  \multicolumn{1}{c|}{0.401} &
  - &
  0.665 \\
\multicolumn{1}{c|}{} &
  InvAD &
  \multicolumn{1}{c}{-} &
  \multicolumn{1}{c|}{\textbf{0.959}} &
  - &
  \multicolumn{1}{c|}{0.975} &
  - &
  \multicolumn{1}{c|}{\textbf{0.950}} &
  - &
  \textbf{0.961} &
  - &
  \multicolumn{1}{c|}{\textbf{0.981}} &
  - &
  \multicolumn{1}{c|}{{\ul 0.621}} &
  - &
  \multicolumn{1}{c|}{\textbf{0.612}} &
  - &
  \multicolumn{1}{c|}{{\ul 0.799}} &
  - &
  \multicolumn{1}{c|}{\textbf{0.441}} &
  - &
  \textbf{0.691} \\
\multicolumn{1}{c|}{} &
  InvAD-lite &
  \multicolumn{1}{c}{-} &
  \multicolumn{1}{c|}{0.931} &
  - &
  \multicolumn{1}{c|}{0.974} &
  - &
  \multicolumn{1}{c|}{0.946} &
  - &
  0.950 &
  - &
  \multicolumn{1}{c|}{0.979} &
  - &
  \multicolumn{1}{c|}{0.592} &
  - &
  \multicolumn{1}{c|}{0.597} &
  - &
  \multicolumn{1}{c|}{0.786} &
  - &
  \multicolumn{1}{c|}{0.426} &
  - &
  0.676 \\
\multicolumn{1}{c|}{} &
  MambaAD &
  \multicolumn{1}{c}{-} &
  \multicolumn{1}{c|}{0.921} &
  - &
  \multicolumn{1}{c|}{0.969} &
  - &
  \multicolumn{1}{c|}{0.938} &
  - &
  0.942 &
  - &
  \multicolumn{1}{c|}{0.976} &
  - &
  \multicolumn{1}{c|}{0.521} &
  - &
  \multicolumn{1}{c|}{0.556} &
  - &
  \multicolumn{1}{c|}{0.777} &
  - &
  \multicolumn{1}{c|}{0.386} &
  - &
  0.643 \\
\multicolumn{1}{c|}{\multirow{-7}{*}{Multi-class}} &
  HierCore (Ours) &
  \multicolumn{1}{c}{-} &
  \multicolumn{1}{c|}{0.941} &
  - &
  \multicolumn{1}{c|}{{\ul 0.978}} &
  - &
  \multicolumn{1}{c|}{0.938} &
  - &
  0.952 &
  - &
  \multicolumn{1}{c|}{0.972} &
  - &
  \multicolumn{1}{c|}{0.580} &
  - &
  \multicolumn{1}{c|}{0.563} &
  - &
  \multicolumn{1}{c|}{0.753} &
  - &
  \multicolumn{1}{c|}{0.393} &
  - &
  0.652 \\ \bottomrule
\end{tabular}
\end{adjustbox}
\end{table}

\FloatBarrier

\section{Evaluation of Requirements for Multi-Class Unsupervised Image Anomaly Detection}
\label{appendix:u_to_uk}

\begin{table}[H]
\caption{Image- and pixel-level anomaly detection performance on the MVTecAD dataset based on F1-score with optimal thresholding. Models are trained under an unknown classes setting. During evaluation, F1-scores for known classes are computed using class-specific optimal thresholds, while a single optimal threshold is used for unknown-class evaluation. Diff. Ratio indicates the ratio of F1-scores under the unknown-class evaluation to those under the known-class evaluation.}\label{tab:exp_u_to_uk_mvtecad}
\vspace{5pt}
\centering
\begin{adjustbox}{max width=\textwidth}
\begin{tabular}{@{}c|c|cccccc|cccccc@{}}
\toprule
\multirow{2}{*}{Categories} &
  \multirow{2}{*}{Evaluation} &
  \multicolumn{6}{c|}{Image-level} &
  \multicolumn{6}{c}{PIxel-level} \\ \cmidrule(l){3-14} 
                            &         & UniAD & ViTAD & InvAD & MambaAD & PatchCore & HierCore & UniAD & ViTAD & InvAD & MambaAD & PatchCore & HierCore \\ \midrule
\multirow{2}{*}{Bottle}     & Known   & 94.6  & 98.4  & 99.2  & 99.2    & 99.2      & 99.2     & 71.0  & 76.5  & 73.7  & 76.5    & 79.4      & 79.5     \\
                            & Unknown & 94.7  & 99.2  & 100.0 & 100.0   & 99.2      & 99.2     & 68.5  & 74.1  & 73.6  & 76.4    & 79.4      & 79.5     \\ \midrule
\multirow{2}{*}{Cable}      & Known   & 88.1  & 93.0  & 94.6  & 95.7    & 97.3      & 96.3     & 51.2  & 47.5  & 53.1  & 47.9    & 57.5      & 64.7     \\
                            & Unknown & 80.3  & 78.0  & 76.0  & 81.1    & 76.0      & 96.3     & 47.6  & 45.1  & 52.9  & 45.3    & 52.3      & 65.4     \\ \midrule
\multirow{2}{*}{Capsule}    & Known   & 90.4  & 94.3  & 95.5  & 93.4    & 97.3      & 97.7     & 36.4  & 48.3  & 50.7  & 44.7    & 56.5      & 56.0     \\
                            & Unknown & 86.7  & 84.0  & 80.2  & 58.4    & 84.0      & 97.7     & 35.8  & 38.8  & 41.1  & 37.9    & 45.9      & 56.0     \\ \midrule
\multirow{2}{*}{Hazelnut}   & Known   & 93.7  & 98.6  & 99.3  & 98.6    & 99.3      & 99.3     & 56.2  & 65.3  & 61.4  & 66.0    & 68.2      & 68.2     \\
                            & Unknown & 81.4  & 77.8  & 85.9  & 88.6    & 78.7      & 99.3     & 50.7  & 51.2  & 55.6  & 54.4    & 57.7      & 68.2     \\ \midrule
\multirow{2}{*}{Metal nut}  & Known   & 96.3  & 97.9  & 99.5  & 98.4    & 98.9      & 98.9     & 67.0  & 77.6  & 81.4  & 80.0    & 87.0      & 87.1     \\
                            & Unknown & 96.7  & 93.9  & 96.4  & 97.9    & 93.5      & 98.9     & 55.4  & 67.8  & 80.3  & 75.4    & 87.0      & 87.1     \\ \midrule
\multirow{2}{*}{Pill}       & Known   & 91.2  & 96.0  & 97.1  & 94.4    & 96.1      & 96.5     & 26.7  & 75.4  & 69.1  & 61.0    & 74.4      & 74.3     \\
                            & Unknown & 91.6  & 96.0  & 96.3  & 94.3    & 96.0      & 96.5     & 18.0  & 61.3  & 43.0  & 52.7    & 64.6      & 74.3     \\ \midrule
\multirow{2}{*}{Screw}      & Known   & 89.3  & 91.1  & 95.1  & 91.5    & 96.7      & 97.1     & 29.4  & 39.2  & 47.4  & 45.6    & 52.1      & 52.0     \\
                            & Unknown & 87.3  & 89.8  & 80.2  & 71.0    & 77.3      & 97.1     & 29.2  & 34.8  & 38.6  & 43.9    & 38.1      & 52.0     \\ \midrule
\multirow{2}{*}{Toothbrush} & Known   & 84.1  & 95.1  & 93.5  & 93.5    & 96.6      & 96.6     & 50.6  & 61.0  & 59.6  & 60.2    & 66.0      & 65.4     \\
                            & Unknown & 85.3  & 92.3  & 95.2  & 95.2    & 96.7      & 96.6     & 49.1  & 52.7  & 51.6  & 53.6    & 63.9      & 65.4     \\ \midrule
\multirow{2}{*}{Transistor} & Known   & 82.5  & 86.0  & 91.6  & 94.7    & 98.7      & 98.7     & 68.0  & 55.1  & 63.2  & 60.7    & 61.1      & 61.3     \\
                            & Unknown & 64.0  & 69.6  & 72.1  & 80.0    & 88.9      & 98.7     & 68.0  & 48.1  & 58.0  & 59.2    & 53.8      & 61.3     \\ \midrule
\multirow{2}{*}{Zipper}     & Known   & 91.3  & 96.6  & 98.3  & 96.7    & 98.3      & 98.3     & 41.4  & 51.0  & 57.0  & 58.8    & 71.4      & 71.7     \\
                            & Unknown & 85.6  & 82.5  & 93.8  & 86.9    & 94.2      & 98.3     & 31.1  & 24.7  & 32.7  & 39.2    & 40.4      & 71.7     \\ \midrule
\multirow{2}{*}{Carpet}     & Known   & 97.7  & 98.9  & 96.5  & 98.9    & 96.0      & 96.0     & 55.7  & 64.5  & 60.9  & 64.8    & 67.8      & 67.8     \\
                            & Unknown & 94.2  & 99.4  & 95.6  & 99.4    & 96.0      & 96.0     & 52.7  & 62.5  & 59.9  & 62.9    & 65.9      & 67.8     \\ \midrule
\multirow{2}{*}{Grid}       & Known   & 93.8  & 98.2  & 96.6  & 99.1    & 98.2      & 99.1     & 34.1  & 37.0  & 46.2  & 47.9    & 53.9      & 54.5     \\
                            & Unknown & 89.1  & 94.4  & 96.5  & 94.4    & 99.1      & 99.1     & 25.0  & 35.2  & 46.1  & 47.9    & 52.8      & 54.5     \\ \midrule
\multirow{2}{*}{Leather}    & Known   & 99.5  & 99.5  & 99.5  & 99.5    & 99.5      & 99.5     & 43.7  & 57.4  & 52.2  & 49.1    & 55.8      & 55.4     \\
                            & Unknown & 100.0 & 100.0 & 98.9  & 100.0   & 100.0     & 99.5     & 31.3  & 33.3  & 36.1  & 31.7    & 51.4      & 55.4     \\ \midrule
\multirow{2}{*}{Tile}       & Known   & 90.7  & 99.4  & 99.4  & 93.8    & 98.2      & 98.2     & 47.6  & 69.1  & 61.8  & 52.1    & 70.9      & 71.0     \\
                            & Unknown & 83.6  & 88.4  & 88.9  & 85.3    & 98.8      & 98.2     & 47.5  & 56.9  & 59.9  & 50.4    & 69.5      & 71.0     \\ \midrule
\multirow{2}{*}{Wood}       & Known   & 94.9  & 96.6  & 97.5  & 95.9    & 96.7      & 96.7     & 45.6  & 58.6  & 52.1  & 48.7    & 58.5      & 58.3     \\
                            & Unknown & 89.6  & 92.3  & 94.5  & 93.8    & 96.7      & 96.7     & 45.4  & 49.1  & 47.0  & 41.2    & 58.1      & 58.3     \\ \midrule
\multirow{3}{*}{Average} &
  Known &
  91.9 &
  96.0 &
  96.9 &
  96.2 &
  {\ul 97.8} &
  \textbf{97.9} &
  48.3 &
  58.9 &
  59.3 &
  57.6 &
  {\ul 65.4} &
  \textbf{65.8} \\
 &
  Unknown &
  87.3 &
  89.2 &
  90.0 &
  88.4 &
  {\ul 91.7} &
  \textbf{97.9} &
  43.7 &
  49.0 &
  51.8 &
  51.5 &
  {\ul 58.7} &
  \textbf{65.9} \\ \cmidrule(l){2-14} 
 &
  Diff.   Ratio &
  {\ul 95.1\%} &
  92.9\% &
  92.9\% &
  91.9\% &
  93.8\% &
  \textbf{100.0\%} &
  {\ul 90.4\%} &
  83.2\% &
  87.2\% &
  89.4\% &
  89.8\% &
  \textbf{100.1\%} \\ \bottomrule
\end{tabular}
\end{adjustbox}
\end{table}

\begin{table}[H]
\caption{Image- and pixel-level anomaly detection performance on the VisA dataset based on F1-score with optimal thresholding. Models are trained under an unknown classes setting. During evaluation, F1-scores for known classes are computed using class-specific optimal thresholds, while a single optimal threshold is used for unknown-class evaluation. Diff. Ratio indicates the ratio of F1-scores under the unknown-class evaluation to those under the known-class evaluation.}\label{tab:exp_u_to_uk_visa}
\vspace{5pt}
\centering
\begin{adjustbox}{max width=\textwidth}
\begin{tabular}{@{}c|c|cccccc|cccccc@{}}
\toprule
\multirow{2}{*}{Categories} &
  \multirow{2}{*}{Evaluation} &
  \multicolumn{6}{c|}{Image-level} &
  \multicolumn{6}{c}{PIxel-level} \\ \cmidrule(l){3-14} 
                             &         & UniAD & ViTAD & InvAD & MambaAD & PatchCore & HierCore & UniAD & ViTAD & InvAD & MambaAD & PatchCore & HierCore \\ \midrule
\multirow{2}{*}{PCB1}        & Known   & 90.2  & 91.2  & 92.6  & 89.9    & 97.0      & 96.5     & 59.6  & 62.1  & 76.4  & 71.3    & 81.1      & 75.9     \\
                             & Unknown & 88.0  & 84.4  & 91.1  & 90.4    & 96.6      & 96.5     & 56.0  & 61.9  & 70.4  & 71.2    & 76.5      & 76.6     \\ \midrule
\multirow{2}{*}{PCB2}        & Known   & 82.9  & 84.3  & 92.3  & 86.3    & 93.5      & 93.0     & 17.0  & 21.2  & 24.0  & 22.6    & 37.7      & 36.1     \\
                             & Unknown & 80.3  & 81.2  & 90.0  & 85.7    & 93.6      & 93.0     & 16.9  & 21.0  & 23.5  & 22.6    & 32.6      & 35.5     \\ \midrule
\multirow{2}{*}{PCB3}        & Known   & 78.3  & 83.5  & 91.8  & 86.4    & 93.5      & 92.9     & 24.1  & 26.4  & 28.4  & 26.8    & 43.9      & 44.0     \\
                             & Unknown & 76.7  & 83.6  & 90.6  & 87.0    & 89.1      & 92.9     & 20.3  & 26.0  & 27.7  & 26.6    & 39.0      & 44.0     \\ \midrule
\multirow{2}{*}{PCB4}        & Known   & 93.1  & 95.6  & 97.4  & 96.4    & 97.5      & 97.5     & 33.6  & 48.3  & 45.2  & 46.5    & 49.1      & 49.1     \\
                             & Unknown & 88.4  & 89.7  & 95.2  & 94.8    & 95.2      & 97.5     & 28.4  & 48.3  & 44.7  & 46.5    & 48.9      & 49.1     \\ \midrule
\multirow{2}{*}{Macaroni1}   & Known   & 75.1  & 74.9  & 87.4  & 80.6    & 95.1      & 94.2     & 16.2  & 19.4  & 29.8  & 24.7    & 39.6      & 39.1     \\
                             & Unknown & 67.1  & 65.7  & 70.4  & 69.0    & 88.8      & 94.2     & 16.2  & 12.4  & 28.5  & 22.9    & 36.0      & 39.1     \\ \midrule
\multirow{2}{*}{Macaroni2}   & Known   & 67.0  & 74.0  & 80.0  & 76.8    & 71.5      & 70.3     & 12.4  & 10.4  & 18.7  & 17.8    & 32.2      & 32.1     \\
                             & Unknown & 64.2  & 67.4  & 69.1  & 71.9    & 71.2      & 70.3     & 11.0  & 2.6   & 17.6  & 17.1    & 31.9      & 32.1     \\ \midrule
\multirow{2}{*}{Capsules}    & Known   & 77.0  & 79.8  & 87.7  & 88.6    & 82.1      & 81.6     & 44.8  & 41.4  & 65.2  & 60.3    & 66.5      & 67.7     \\
                             & Unknown & 77.5  & 76.9  & 87.2  & 87.3    & 73.5      & 81.6     & 41.8  & 39.4  & 59.0  & 51.2    & 50.3      & 67.7     \\ \midrule
\multirow{2}{*}{Candles}     & Known   & 87.9  & 85.0  & 90.1  & 91.5    & 95.9      & 96.5     & 32.8  & 26.6  & 34.5  & 31.0    & 43.2      & 43.6     \\
                             & Unknown & 87.3  & 79.6  & 77.7  & 90.7    & 88.9      & 96.5     & 32.5  & 16.1  & 29.1  & 30.8    & 38.2      & 43.6     \\ \midrule
\multirow{2}{*}{Cashew}      & Known   & 91.0  & 85.2  & 94.4  & 88.8    & 95.5      & 95.5     & 59.4  & 62.9  & 61.5  & 53.0    & 61.4      & 61.2     \\
                             & Unknown & 84.6  & 84.5  & 89.0  & 89.3    & 94.5      & 95.5     & 57.2  & 57.8  & 55.3  & 49.3    & 54.6      & 61.2     \\ \midrule
\multirow{2}{*}{Chewing gum} & Image   & 94.9  & 92.6  & 95.3  & 94.2    & 97.5      & 97.5     & 58.4  & 59.0  & 62.6  & 59.8    & 58.8      & 51.1     \\
                             & Pixel   & 89.4  & 79.0  & 91.2  & 79.0    & 98.0      & 97.5     & 44.0  & 32.5  & 49.1  & 42.0    & 44.0      & 51.1     \\ \midrule
\multirow{2}{*}{Fryum}       & Image   & 84.7  & 90.2  & 91.8  & 88.7    & 91.9      & 92.1     & 53.1  & 51.1  & 53.1  & 52.6    & 48.7      & 48.6     \\
                             & Pixel   & 83.4  & 90.2  & 88.5  & 85.9    & 91.7      & 92.1     & 41.6  & 28.3  & 37.8  & 35.5    & 36.7      & 48.6     \\ \midrule
\multirow{2}{*}{Pipe fryum}  & Image   & 93.4  & 94.9  & 97.5  & 96.1    & 99.0      & 98.5     & 59.7  & 66.7  & 66.4  & 59.2    & 61.9      & 61.9     \\
                             & Pixel   & 90.7  & 89.2  & 95.8  & 94.4    & 98.5      & 98.5     & 58.2  & 66.5  & 65.0  & 56.8    & 58.7      & 61.9     \\ \midrule
\multirow{3}{*}{Average} &
  Known &
  84.6 &
  85.9 &
  91.5 &
  88.7 &
  \textbf{92.5} &
  {\ul 92.2} &
  39.3 &
  41.3 &
  47.2 &
  43.8 &
  \textbf{52.0} &
  {\ul 50.9} \\
 &
  Unknown &
  81.5 &
  81.0 &
  86.3 &
  85.5 &
  {\ul 90.0} &
  \textbf{92.2} &
  35.3 &
  34.4 &
  42.3 &
  39.4 &
  {\ul 45.6} &
  \textbf{50.9} \\ \cmidrule(l){2-14} 
 &
  Diff.   Ratio &
  96.3\% &
  94.2\% &
  94.3\% &
  96.4\% &
  {\ul 97.2\%} &
  \textbf{100.0\%} &
  {\ul 90.0\%} &
  83.3\% &
  89.7\% &
  90.0\% &
  87.8\% &
  \textbf{100.0\%} \\ \bottomrule
\end{tabular}
\end{adjustbox}
\end{table}

\begin{table}[H]
\caption{Image- and pixel-level anomaly detection performance on the MPDD dataset based on F1-score with optimal thresholding. Models are trained under an unknown classes setting. During evaluation, F1-scores for known classes are computed using class-specific optimal thresholds, while a single optimal threshold is used for unknown-class evaluation. Diff. Ratio indicates the ratio of F1-scores under the unknown-class evaluation to those under the known-class evaluation.}\label{tab:exp_u_to_uk_mpdd}
\vspace{5pt}
\centering
\begin{adjustbox}{max width=\textwidth}
\begin{tabular}{@{}c|c|cccccc|cccccc@{}}
\toprule
\multirow{2}{*}{Categories} &
  \multirow{2}{*}{Evaluation} &
  \multicolumn{6}{c|}{Image-level} &
  \multicolumn{6}{c}{PIxel-level} \\ \cmidrule(l){3-14} 
 &
   &
  UniAD &
  ViTAD &
  InvAD &
  MambaAD &
  PatchCore &
  HierCore &
  UniAD &
  ViTAD &
  InvAD &
  MambaAD &
  PatchCore &
  HierCore \\ \midrule
\multirow{2}{*}{\begin{tabular}[c]{@{}c@{}}Bracket\\ Black\end{tabular}} &
  Known &
  78.9 &
  80.4 &
  80.4 &
  80.0 &
  84.1 &
  83.9 &
  1.6 &
  9.6 &
  9.6 &
  16.8 &
  28.7 &
  28.4 \\
 &
  Unknown &
  78.5 &
  81.4 &
  81.4 &
  78.6 &
  76.5 &
  83.5 &
  1.4 &
  0.0 &
  0.0 &
  13.9 &
  7.5 &
  30.2 \\ \midrule
\multirow{2}{*}{\begin{tabular}[c]{@{}c@{}}Bracket\\ Brown\end{tabular}} & Known & 92.6 & 90.1 & 90.1 & 93.5 & 95.1 & 94.3 & 32.1 & 23.4 & 23.4 & 25.5 & 32.6 & 31.3 \\
 &
  Unknown &
  92.6 &
  90.1 &
  90.1 &
  85.1 &
  88.4 &
  94.2 &
  17.0 &
  0.0 &
  0.0 &
  15.4 &
  4.7 &
  31.0 \\ \midrule
\multirow{2}{*}{\begin{tabular}[c]{@{}c@{}}Bracket\\ White\end{tabular}} &
  Known &
  71.8 &
  73.3 &
  73.3 &
  87.7 &
  86.8 &
  85.2 &
  1.1 &
  2.2 &
  2.2 &
  22.6 &
  25.7 &
  26.3 \\
 &
  Unknown &
  67.4 &
  66.7 &
  66.7 &
  88.1 &
  86.8 &
  85.2 &
  0.8 &
  0.0 &
  0.0 &
  17.5 &
  0.6 &
  26.3 \\ \midrule
\multirow{2}{*}{Connector} &
  Image &
  63.4 &
  82.8 &
  82.8 &
  92.9 &
  96.3 &
  96.3 &
  15.5 &
  48.5 &
  48.5 &
  55.8 &
  63.4 &
  62.1 \\
 &
  Pixel &
  57.1 &
  58.3 &
  58.3 &
  84.8 &
  82.4 &
  96.3 &
  10.9 &
  3.3 &
  3.3 &
  53.1 &
  6.2 &
  62.1 \\ \midrule
\multirow{2}{*}{Metal Plate} &
  Image &
  83.8 &
  99.3 &
  99.3 &
  99.3 &
  99.3 &
  99.3 &
  57.7 &
  88.0 &
  88.0 &
  87.2 &
  86.9 &
  86.9 \\
 &
  Pixel &
  84.5 &
  84.5 &
  84.5 &
  86.1 &
  87.7 &
  99.3 &
  57.2 &
  88.0 &
  88.0 &
  85.6 &
  86.8 &
  86.9 \\ \midrule
\multirow{2}{*}{Tubes} &
  Image &
  82.4 &
  83.5 &
  83.5 &
  93.9 &
  91.6 &
  91.7 &
  14.8 &
  54.5 &
  54.5 &
  69.4 &
  66.5 &
  66.8 \\
 &
  Pixel &
  81.2 &
  81.2 &
  81.2 &
  81.2 &
  81.2 &
  91.7 &
  5.2 &
  43.6 &
  43.6 &
  57.3 &
  66.1 &
  66.8 \\ \midrule
\multirow{3}{*}{Average} &
  Known &
  78.8 &
  84.9 &
  84.9 &
  91.2 &
  \textbf{92.2} &
  {\ul 91.8} &
  20.5 &
  37.7 &
  37.7 &
  46.2 &
  \textbf{50.6} &
  {\ul 50.3} \\
 &
  Unknown &
  76.9 &
  77.0 &
  77.0 &
  {\ul 84.0} &
  83.8 &
  \textbf{91.7} &
  15.4 &
  22.5 &
  22.5 &
  {\ul 40.5} &
  28.6 &
  \textbf{50.5} \\ \cmidrule(l){2-14} 
 &
  Diff. Ratio &
  {\ul 97.5\%} &
  90.8\% &
  90.8\% &
  92.1\% &
  90.9\% &
  \textbf{99.9\%} &
  75.4\% &
  59.6\% &
  59.6\% &
  {\ul 87.5\%} &
  56.5\% &
  \textbf{100.5\%} \\ \bottomrule
\end{tabular}
\end{adjustbox}
\end{table}

\begin{table}[H]
\caption{Image- and pixel-level anomaly detection performance on the BTAD dataset based on F1-score with optimal thresholding. Models are trained under an unknown classes setting. During evaluation, F1-scores for known classes are computed using class-specific optimal thresholds, while a single optimal threshold is used for unknown-class evaluation. Diff. Ratio indicates the ratio of F1-scores under the unknown-class evaluation to those under the known-class evaluation.}\label{tab:exp_u_to_uk_btad}
\vspace{5pt}
\centering
\begin{adjustbox}{max width=\textwidth}
\begin{tabular}{@{}c|c|cccccc|cccccc@{}}
\toprule
\multirow{2}{*}{Categories} &
  \multirow{2}{*}{Evaluation} &
  \multicolumn{6}{c|}{Image-level} &
  \multicolumn{6}{c}{PIxel-level} \\ \cmidrule(l){3-14} 
 &
   &
  UniAD &
  ViTAD &
  InvAD &
  MambaAD &
  PatchCore &
  HierCore &
  UniAD &
  ViTAD &
  InvAD &
  MambaAD &
  PatchCore &
  HierCore \\ \midrule
\multirow{2}{*}{01} &
  Image &
  96.8 &
  94.8 &
  97.9 &
  93.6 &
  97.0 &
  96.9 &
  58.5 &
  55.7 &
  62.2 &
  59.7 &
  59.3 &
  59.6 \\
 &
  Pixel &
  93.5 &
  89.9 &
  96.8 &
  94.6 &
  86.0 &
  96.9 &
  56.9 &
  0.4 &
  46.5 &
  41.4 &
  55.9 &
  59.6 \\ \midrule
\multirow{2}{*}{02} &
  Image &
  92.7 &
  93.1 &
  93.9 &
  92.5 &
  93.2 &
  93.0 &
  51.8 &
  67.6 &
  62.4 &
  58.3 &
  60.3 &
  60.2 \\
 &
  Pixel &
  83.4 &
  90.4 &
  79.5 &
  82.2 &
  64.2 &
  93.0 &
  51.8 &
  67.4 &
  62.1 &
  58.0 &
  60.2 &
  60.2 \\ \midrule
\multirow{2}{*}{03} &
  Image &
  89.7 &
  90.6 &
  90.0 &
  92.1 &
  90.9 &
  88.5 &
  52.1 &
  47.5 &
  59.5 &
  49.0 &
  49.8 &
  49.7 \\
 &
  Pixel &
  66.0 &
  53.4 &
  64.6 &
  50.4 &
  84.9 &
  88.5 &
  50.5 &
  0.4 &
  51.2 &
  48.6 &
  49.4 &
  49.7 \\ \midrule
\multirow{3}{*}{Average} &
  Known &
  93.1 &
  92.9 &
  \textbf{93.9} &
  92.7 &
  {\ul 93.7} &
  92.8 &
  54.1 &
  {\ul 56.9} &
  \textbf{61.4} &
  55.7 &
  56.5 &
  56.5 \\
 &
  Unknown &
  {\ul 81.0} &
  77.9 &
  80.3 &
  75.7 &
  78.4 &
  \textbf{92.8} &
  53.0 &
  22.8 &
  53.3 &
  49.3 &
  {\ul 55.2} &
  \textbf{56.5} \\ \cmidrule(l){2-14} 
 &
  Diff. Ratio &
  {\ul 87.0\%} &
  83.9\% &
  85.5\% &
  81.7\% &
  83.7\% &
  \textbf{100.0\%} &
  {\ul 98.0\%} &
  40.0\% &
  86.8\% &
  88.6\% &
  97.7\% &
  \textbf{100.0\%} \\ \bottomrule
\end{tabular}
\end{adjustbox}
\end{table}

\end{document}